\documentclass{article}

\usepackage{arxiv}

\usepackage[utf8]{inputenc} % allow utf-8 input
\usepackage[T1]{fontenc}    % use 8-bit T1 fonts
\usepackage{hyperref}       % hyperlinks
\usepackage{url}            % simple URL typesetting
\usepackage{booktabs}       % professional-quality tables
\usepackage{amsfonts}       % blackboard math symbols
\usepackage{nicefrac}       % compact symbols for 1/2, etc.
\usepackage{microtype}      % microtypography
\usepackage{lipsum}
\usepackage{graphicx}
\graphicspath{ {./images/} }

\usepackage{multirow}
\usepackage{makecell}
\usepackage{float}
\usepackage{placeins}

\usepackage{tabularx}
\usepackage{siunitx}
\usepackage{caption}
\usepackage{subcaption}

\usepackage{pifont}% http://ctan.org/pkg/pifont

\title{Dynamic Cluster Data Sampling for Efficient and Long-Tail-Aware Vision-Language Pre-training}

\author{Mingliang Liang$^{1}$, Zhuoran Liu$^{2}$\thanks{Corresponding author. Work was done while at Radboud University.}, Arjen P. de Vries$^{1}$, Martha Larson$^{1}$ \\
$^{1}$Institute for Computing and Information Sciences, Radboud University, Nijmegen, the Netherlands \\
$^{2}$University of Amsterdam, Amsterdam, the Netherlands \\
{\tt\small \{mliang, z.liu, a.devries, mlarson\}@cs.ru.nl}
}

\begin{document}
\maketitle

\begin{abstract}
The computational cost of training a vision-language model (VLM) can be reduced by sampling the training data.
Previous work on efficient VLM pre-training has pointed to the importance of semantic data balance, adjusting the distribution of topics in the data to improve VLM accuracy.
However, existing efficient pre-training approaches may disproportionately remove rare concepts from the training corpus. 
As a result, \emph{long-tail concepts} remain insufficiently represented in the training data and are not effectively captured during training.
In this work, we introduce a \emph{dynamic cluster-based sampling approach (DynamiCS)} that downsamples large clusters of data and upsamples small ones.
We first demonstrate the advantage of our cluster-scaling approach, which maintains the relative order of semantic clusters in the data and emphasizes the long-tail.
This approach contrasts with current work, which focuses only on flattening the semantic distribution of the data.
Then, we show the importance of dynamic sampling, which applies sampling at each epoch to improve cross-epoch data diversity and make upsampling practical.
Our experiments show that DynamiCS reduces the computational cost of VLM training and provides a performance advantage for long-tail concepts.
Code available at \url{https://github.com/MingliangLiang3/DynamiCS}.
\end{abstract}

% keywords can be removed

\section{Introduction}
\label{sec:intro}

Vision-Language Models (VLMs) demonstrate strong transferability~\cite{kaplan2020scalinglaw,radford2021clip,cherti2023openclip,scaling2021Jia,desai2021virtex,zhang2022contrastive,li2023flip,zhai2023siglip}. 
Pre-trained VLMs can be applied to classification and image-text retrieval tasks, 
and their image encoders are widely used in Multimodal Large Language Models (MLLM) and generative models~\cite{li2022blip,liu2023llava,openai2024gpt4,ramesh2022dalle}.
CLIP~\cite{radford2021clip}, one of the popular VLMs, is trained on extremely large-scale datasets, requiring substantial GPU resources for pre-training. 
The training costs of CLIP have given rise to \emph{cost-saving} training approaches, notably, RECLIP~\cite{li2023reclip}, FLIP~\cite{li2023flip}, and CLIPA~\cite{li2023clipa}, which maintain VLM performance but reduce training costs by reducing the amount of text or image information used from each sample for training.

CLIP's large-scale pre-training data is collected from the web, but the data is curated to improve \emph{semantic data balance}, i.e., to adjust the distribution of topics in the data, which in the wild typically has a very fat head and a very long tail.
In~\cite{xu2024metaclip}, the advantages of CLIP data curation are described as avoiding biases and balancing the data over the metadata, which corresponds to semantic categories. 
These advantages motivate the data curation approach in MetaCLIP~\cite{xu2024metaclip}, which limits the number of data samples associated with each metadata category to 20k, effectively chopping off the fat head of the distribution.

\begin{figure}[!t]
    \centering
    \includegraphics[width=1.0\linewidth
    ]{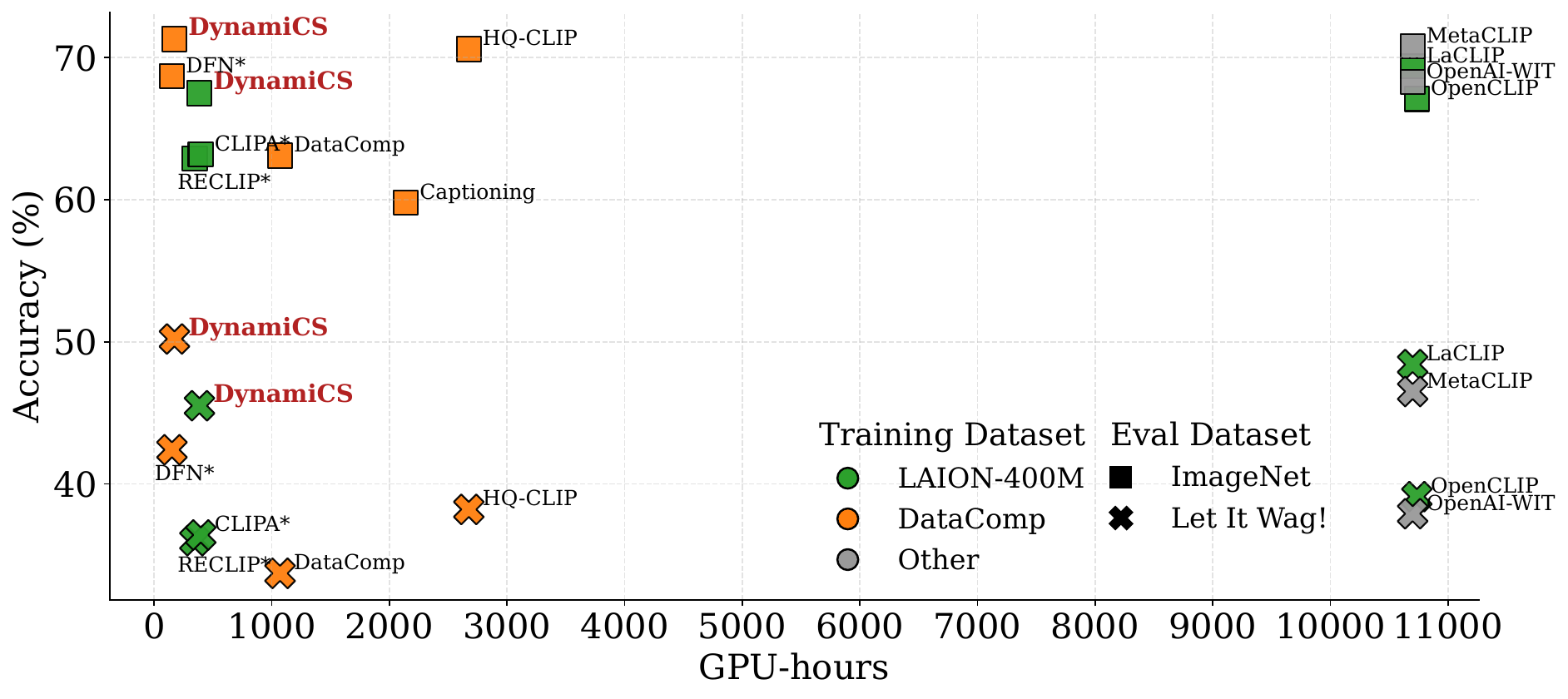}
    \caption{
    \textbf{Zero-shot top-1 accuracy on ImageNet-1K~\cite{deng2009imagenet} and \textit{Let It Wag!}~\cite{udandarao2024no} (long-tail test set).}
    DynamiCS outperforms cost-saving baselines (RECLIP~\cite{li2023reclip}, FLIP~\cite{li2023flip}, CLIPA~\cite{li2023clipa}) and dual-purpose approaches (DataComp~\cite{gadre2023datacomp}, DFN~\cite{fang2023dfn}, Captioning~\cite{nguyen2023improving}) while using less computational resources and achieves accuracy competitive with full-scale pre-training, e.g., OpenCLIP~\cite{cherti2023openclip}.
    Experiments are conducted with ViT-B/16 pre-training of 6 epochs, applying different strategies to LAION-400M~\cite{schuhmann2021laion400m} or DataComp~\cite{gadre2023datacomp}. 
     }
    \label{fig:accuracy_trade_off}

\end{figure}

\emph{Dual-purpose approaches} apply data sampling for the simultaneous purposes of training cost reduction and achieving semantic balance.
This gap is left open by MetaCLIP~\cite{xu2024metaclip}, whose goal is a semantically balanced training set.
Metadata is not the only means of identifying semantic structure, rather clustering can be used instead, as with density-based pruning (DBP)~\cite{abbas2023semdedupdbp}.
Dual-purpose approaches can also combine data sampling for training cost reduction with sampling to improve data quality. 
For example, approaches have removed text-image pairs using another already-trained CLIP as a filter~\cite{gadre2023datacomp,fang2023dfn} and integrating information from automatically generated captions~\cite{nguyen2023improving}.
However, these approaches complicate the goal of training cost reduction since they all make critical use of another already-trained VLM. 

In this paper, we propose a dual-purpose approach for VLM training reduction, DynamiCS,
which reduces costs with dynamic sampling and addresses semantic data balance with cluster scaling.  
In Fig.~\ref{fig:accuracy_trade_off}, we observe that DynamiCS achieves a substantial speedup over OpenCLIP~\cite{cherti2023openclip}, MetaCLIP~\cite{xu2024metaclip}, and HQ-CLIP~\cite{wei2025hq}, while also yielding a slight improvement on ImageNet-1K zero-shot classification.
It improves (top of graph) over other cost-saving training approaches (RECLIP, CLIPA, FLIP) and other dual-purpose approaches (DFN, DataComp, Captioning) in terms of training cost and also performance.
The improvements are particularly remarkable on \emph{long-tail test data} (bottom of the graph), where the pre-trained DynamiCS ViT-B/16 models substantially outperform the baselines on long-tail benchmark \textit{Let It Wag!}~\cite{udandarao2024no} classification task.

The inspiration for DynamiCS is a change of perspective from previous semantic data balancing approaches, which ``aim for even'' distributions of pre-training data over semantic categories.
The ``aim for even'' philosophy is represented by MetaCLIP~\cite{xu2024metaclip}, which chops off the fat head, and DBP~\cite{abbas2023semdedupdbp}, which homogenizes densities with a comparable effect. 
Cluster scaling instantiates an ``aim for utility'' philosophy, which does not necessarily result in a flattened distribution.
Specifically, it downsamples the semantic fat head, without aiming to completely eliminate it. At the same time, it upsamples the semantic long tail, which improves the VLM performance on long-tail concepts.

In sum, this paper makes the following contributions:
\begin{itemize}
    \item We point out that data sampling should involve not only down- but also upsampling, in order to improve VLM performance on the long tail concepts. 
    \item We show that dynamic sampling complements cluster scaling: by drawing a different subset of data at each epoch, it improves cross-epoch diversity and makes upsampling practical without increasing training cost.
    \item On the basis of these insights, we propose a dynamic cluster-based sampling approach (DynamiCS), which outperforms other cost-reducing VLM training approaches and is competitive with full-scale OpenCLIP~\cite{cherti2023openclip} while requiring only about 3\% of its training costs.
\end{itemize}

The paper is structured as follows: in Sec.~\ref{sec:related_work}, we cover related work. 
Then, we introduce DynamiCS (Sec.~\ref{sec:method}) and our experimental setup (Sec.~\ref{sec:setup}). We provide further motivation with experimental analysis on a smaller-scale dataset (Sec.~\ref{sec:analysis}). 
In Sec.~\ref{sec:results}, we present experimental results on a large-scale dataset that demonstrate the advances of DynamiCS. Sec.~\ref{sec:conclusion} provides discussion and outlook.

\section{Related Work}
\label{sec:related_work}

\begin{figure*}[!t]
    \centering
    \begin{subfigure}[b]{0.32\textwidth}
        \centering
        \includegraphics[width=\textwidth]{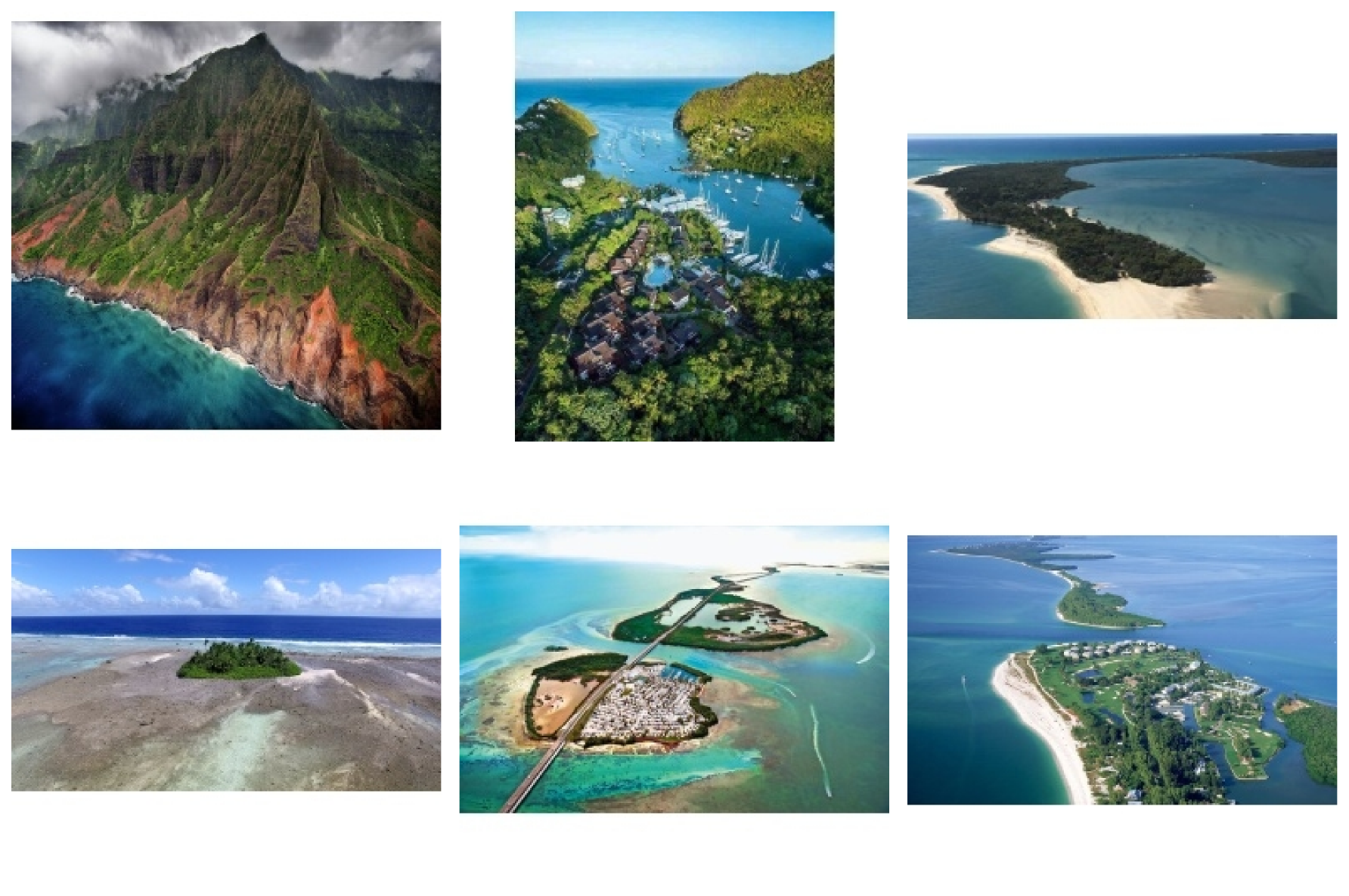}
    \end{subfigure}
    \hfill
    \begin{subfigure}[b]{0.32\textwidth}
        \centering
        \includegraphics[width=\textwidth]{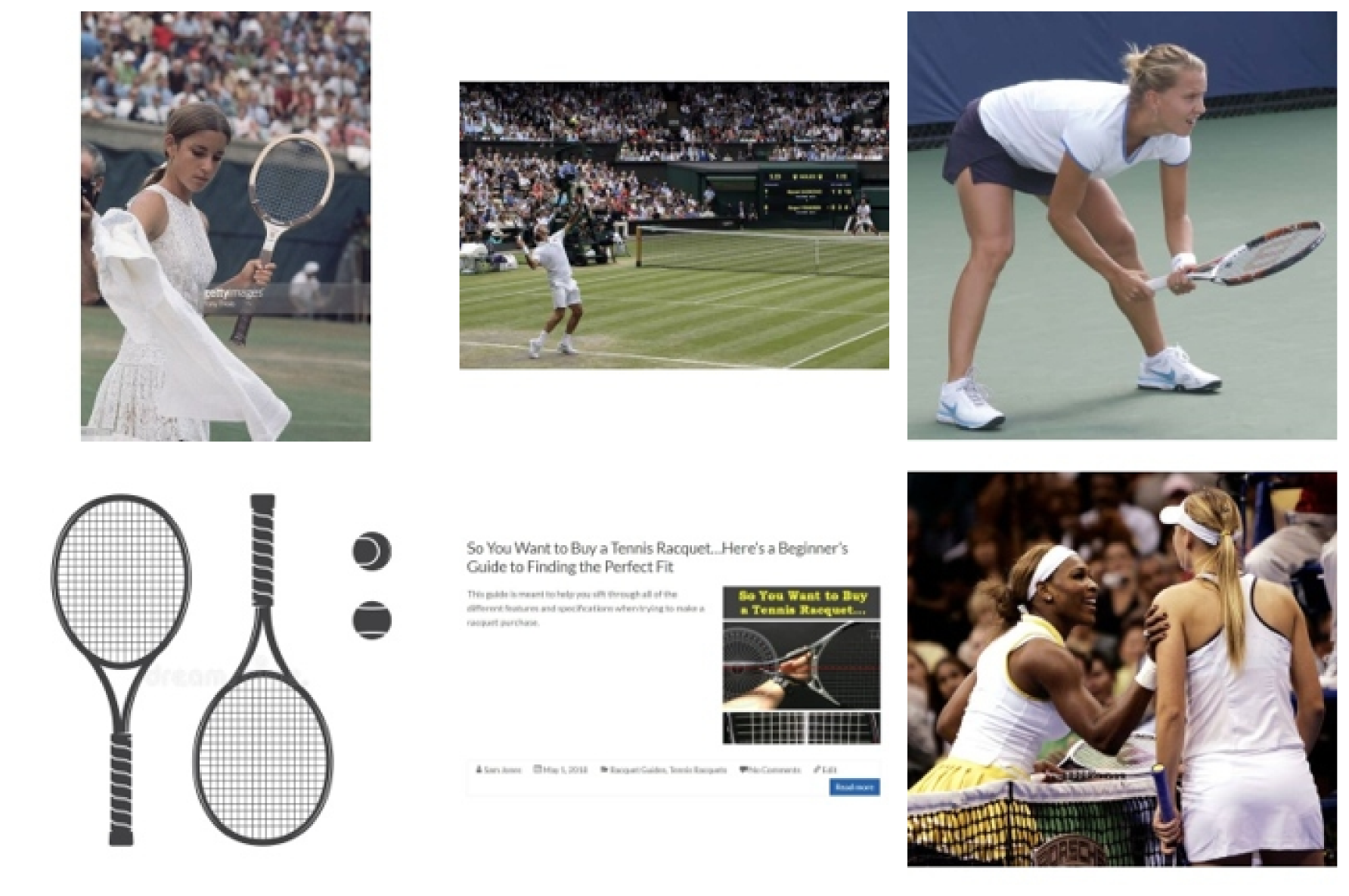}
    \end{subfigure}
    \hfill
    \begin{subfigure}[b]{0.32\textwidth}
        \centering
        \includegraphics[width=\textwidth]{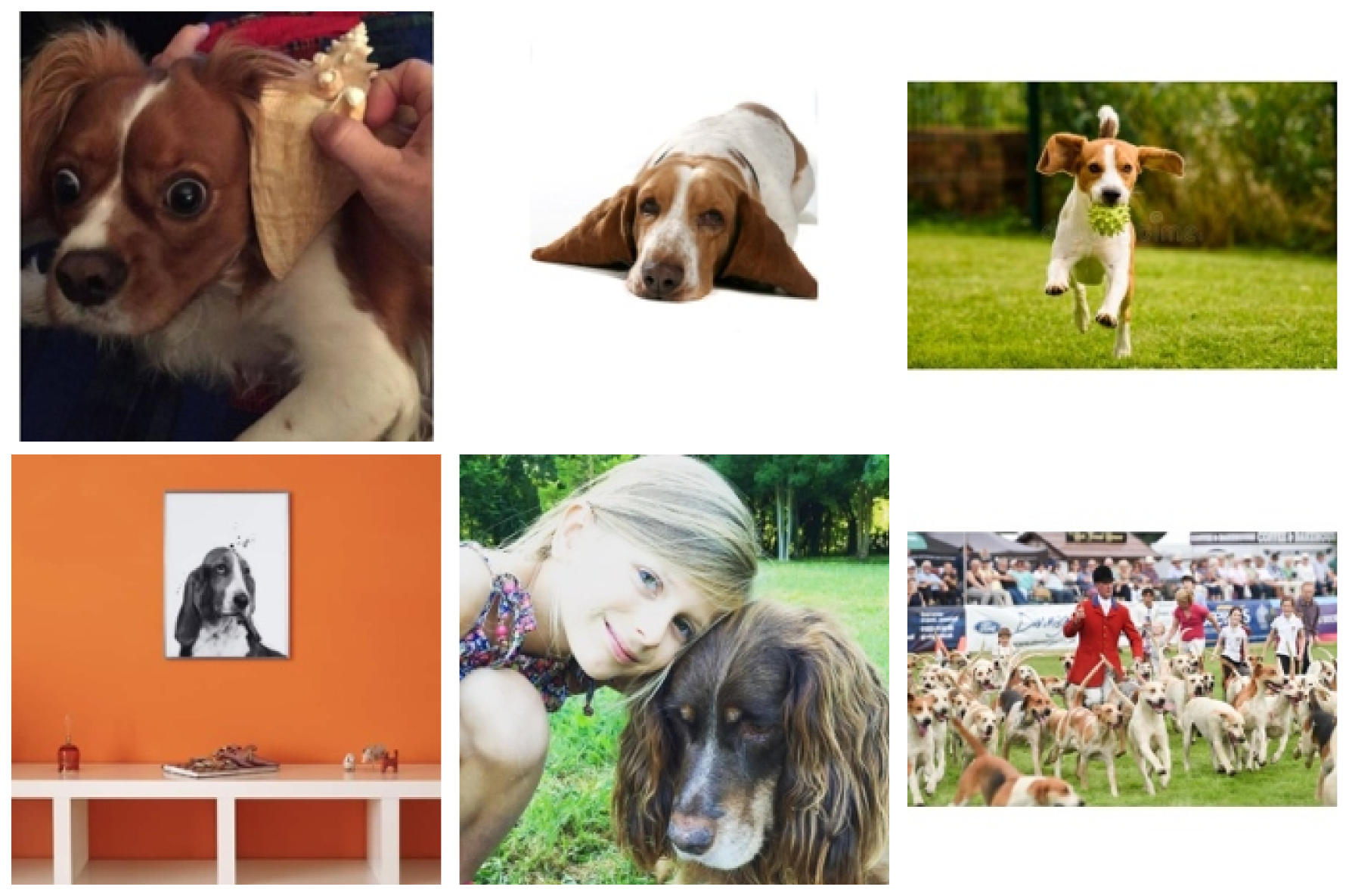}
    \end{subfigure}
        \caption{
        Image examples from three semantic clusters (e.g., ``sea,'' ``tennis'' ``dog''), where visually similar concepts are grouped together.  
    }
    \label{fig:cluster_exa}

\end{figure*}

\subsection{Vision-Language Models}

Large-scale Vision-Language Models (VLMs) have demonstrated remarkable transferability~\cite {kaplan2020scalinglaw,radford2021clip,cherti2023openclip,scaling2021Jia,desai2021virtex,zhang2022contrastive,li2023flip,zhai2023siglip},
which forms the basis for their widespread applicability and effectiveness in downstream tasks.
CLIP (Contrastive Language-Image Pre-Training) exemplifies this paradigm by learning visual–semantic embeddings from language through contrastive learning~\cite{radford2021clip,oord2018InfoNCE}.
CLIP pre-training relies on large-scale datasets and can use millions or billions of image–text pairs.

Following CLIP~\cite{radford2021clip}, subsequent work, such as OpenCLIP~\cite{cherti2023openclip} and
MetaCLIP~\cite{xu2024metaclip}, has advanced the paradigm through open-source training 
and improved data curation strategies.
Reproducible training with publicly accessible and diverse image–text pairs has become a key factor in advancing large-scale VLMs, which facilitate downstream applications.
However training, as mentioned in Sec.~\ref{sec:intro}, requires substantial resources.
In this paper, we focus on reducing the costs of VLM training while preserving model effectiveness.

\subsection{Cost-reducing VLM Training}

Approaches that reduce the training costs of VLMs fall into two categories: cost-saving approaches that reduce the amount of information in each sample and data filtering approaches, which use data sampling to make the training set smaller. We discuss each in turn.

\noindent \textbf{Reducing Image/Text Tokens.} 
RECLIP~\cite{li2023reclip} introduces the use of low-resolution images for CLIP pre-training followed by high-resolution fine-tuning, achieving a substantial reduction in training cost.
Taking the same idea to the patch level, FLIP~\cite{li2023flip} uses random image masking, where only a subset of patches is used for training. 
Similar token number reduction methods based on image-token distribution have been explored with A-CLIP~\cite{yang2023aclip}, GLIP~\cite{mingliang2024glip}, and CLIP-PGS~\cite{pei2025clippgs}, and based on word frequency with CLIPF~\cite{liang2024clipf}. 
CLIPA~\cite{li2023clipa} investigates and combines different token-level strategies, including low-resolution images with truncation, random, block, and syntax-based text masking.
All token-reducing pre-training methods achieve comparable zero-shot performance at substantially lower cost compared to full-scale training~\cite{li2023clipa}.
Reducing the number of tokens has become a common practice in cost-reducing VLM training; however, it does not account for semantic data balance.
In this paper, we propose a dual-purpose approach for reducing VLM training costs while addressing semantic data balance through cluster scaling.

\noindent \textbf{Training Data Filtering.}
The pre-training datasets used for CLIP are large-scale image–text collections crawled from the Internet, such as CC3M~\cite{sharma2018cc3m}, CC12M~\cite{changpinyo2021cc12m}, LAION-400M~\cite{schuhmann2021laion400m}, LAION-2B~\cite{schuhmann2022laionb}, and DataComp~\cite{gadre2023datacomp}.
These datasets are inherently noisy, contain many duplicates or near-duplicates, and are semantically highly imbalanced~\cite{udandarao2024no,wen2024what,parashar2024neglected,xu2024metaclip}.
Several studies have observed that CLIP’s class-wise performance exhibits a log-linear relationship with class frequency~\cite{wen2024what,udandarao2024no,parashar2024neglected}.
Recent studies focus on filtering semantically redundant images out of pre-training datasets~\cite{abbas2023semdedup,webster2023duplication}.
As mentioned in Sec.~\ref{sec:intro}, 
DBP~\cite{abbas2023semdedupdbp} is a dual-purpose approach that simultaneously seeks to reduce training cost and achieve semantic data balance.
It first removes duplicates, then clusters and prunes the data based on a measure of cluster complexity, resulting in consistently better performance than random pruning.
However, while pursuing the aim of uniform data density, it does not explicitly take into account long-tail concepts. 
Instead, long-tail clusters can lose a larger proportion of samples. 

Other dual-purpose approaches combine training-cost reduction with the selection of high-quality data.
DataComp~\cite{gadre2023datacomp} and DFN~\cite{fang2023dfn} use a pre-trained CLIP model to filter misaligned image–text pairs with low similarity scores.
Synthetic captions generated with LLMs or VLLMs can also be used to improve alignment~\cite{wei2025hq,li2025whatif,nguyen2023improving,li2025openvision,fan2023laclip}, however, this may reduce data diversity and ultimately limit model generalization~\cite{nguyen2023improving}.
As noted in Sec.~\ref{sec:intro}, these approaches rely on the existence of another already-trained VLM. 
Overall, long-tail concepts remain underrepresented during VLM pre-training~\cite{udandarao2024no}.

\subsection{Dynamic Data Sampling}
Previous data pruning methods utilize a fixed subset of the dataset during the training~\cite{abbas2023semdedup,abbas2023semdedupdbp,xu2024metaclip}, after which the model only sees part of the training data.
An alternative method chooses a different pruning criterion in different stages of pre-training. 
One example is ClusterClip~\cite{shao2024clusterclip}, which balances the clusters within each batch during large language model training, but it does not aim to reduce the overall training cost.
In our work, we balance the dataset of VLM training by subsampling high-frequency concepts and upsampling low-frequency concepts. 
During pre-training, data samples are dynamically drawn in each training epoch.

\section{DynamiCS}
\label{sec:method}

In this section, we first discuss common issues with current cost-reducing VLM training, and then introduce our dynamic cluster data sampling method.

\subsection{The Challenge of Semantic Data Balance}
\label{sec:Imbalance Semantic Frequency}

Fig.~\ref{fig:cluster_exa} provides an impression of the semantic structure of CLIP training data that is revealed via clustering. 
We generated image embeddings using the pre-trained DINOv2-ViT-B/16~\cite{oquab2024dinov2} model 
and clustered them using K-means~\cite{abbas2023semdedup,abbas2023semdedupdbp}.
Following the definition of ``concept''~\cite{udandarao2024no} that includes object classes, i.e., lemmatized nouns in captions/generation prompts, we see in Fig.~\ref{fig:cluster_exa}, that clusters can be associated with dominant concepts.
For example, one cluster is associated with beach, view, sunset, and sea, which relate to the concept of ``sea''. 
Another cluster includes tennis, court, player, and ball, which are associated with ``tennis''.

However, CLIP training faces the challenge that training data contains a few very large clusters and a large number of small clusters.
As mentioned in Sec.~\ref{sec:intro}, in the wild the distribution of topics in the data has a very fat head and a very long tail. 
Specifically, if we cluster DataComp~\cite{gadre2023datacomp} using image embeddings into 50k clusters, each cluster contains approximately 12k samples on average. 
However, the distribution is too imbalanced.
We observe that some clusters contain more than 1 million points, while others have fewer than 1k. 
Other datasets, such as CC3M~\cite{sharma2018cc3m}, YFCC-15M~\cite{thomee2016yfcc100m}, and the LAION-400M dataset also exhibit an imbalance in concept frequency~\cite{udandarao2024no,abbas2023semdedupdbp}. 
Also, a similar long-tail concept distribution of the pre-training dataset of VLMs has been observed in previous works~\cite{udandarao2024no,parashar2024neglected,wen2024what}.

These observations concerning semantic balance motivate our introduction of a dual-purpose approach that simultaneously reduces training costs and addresses semantic data balance. 
However, we observe that existing methods designed to achieve semantic data balance, represented by MetaCLIP~\cite{xu2024metaclip} and DBP~\cite{abbas2023semdedupdbp}, pursue an ``aim for even'' philosophy, mentioned in Sec.~\ref{sec:intro}, which seeks to flatten the semantic distribution without devoting any specific attention to semantic categories that are not well represented, i.e., comprise the long tail. 
Because DynamiCS reduces training costs with dynamic training, we have the freedom to explore an ``aim for utility'' philosophy that downsamples to reduce, but not completely flatten, the fat head, and upsamples to ensure the long-tail is well represented. 
The ``aim for utility'' approach is supported by recent work that establishes that CLIP training is robust to long-tail distributions~\cite{wen2024what}, meaning that it should not be necessary to aim for a completely flat distribution, and also by work that points to the continuing challenge of the long semantic tail~\cite{parashar2024neglected,udandarao2024no,wen2024what}.

\subsection{Cluster Scaling}

DynamiCS controls semantic balance with Eq.~\ref{equ:shrinkage}:
\begin{equation}
    S_i = \frac{c_i^{\alpha}}{\sum_{j=1}^{N} c_j^{\alpha}} \cdot T
    \label{equ:shrinkage}
\end{equation}
where
$N$ is the total number of clusters,
$c_i$ = original size of cluster $i$,
$\alpha \ge 0$ is a scaling factor that controls how aggressively large clusters are down-sampled and small clusters are up-sampled,
$T$ is the target total number of samples across all clusters,
$S_i$ is the resampled size of cluster $i$.
When $S_i > c_i$ (up-sampled clusters), samples are drawn with replacement.

The motivation for Eq.~\ref{equ:shrinkage} is twofold:
First, it provides the basic scaling we need, reducing the number of samples in large clusters and increasing the number in small clusters, while preserving the relative order of cluster sizes.
Second, it allows us to control the shape of the semantic distribution using $\alpha$, which can range from 0 to a large number.
The case of $\alpha=0$ represents the extreme of the ``aim for even'' philosophy in which all clusters contain the same number of samples.
The case of $\alpha=1$ reduces to random sampling, an important baseline for data sampling.
When $\alpha$ is large, the formula mimics cases in which long-tail semantic clusters are neglected to the point of being eliminated, as happens in~\cite{abbas2023semdedup} under de-duplication thresholds corresponding to aggressive pruning. 

We create the clusters using k-means with cosine similarity to group the image embeddings.
After clustering the embeddings, some clusters remain semantically redundant because their centroids are close to each other in the embedding space.
To reduce this redundancy, we merge centroids whose cosine similarity exceeds a threshold.
The clustering approach is based on the assumption that semantic concepts are well represented in the image embedding space of pre-trained vision models.
Our inspection of the clustering results, cf. Fig.~\ref{fig:cluster_exa} supports this assumption.

We note that Eq.~\ref{equ:shrinkage} is reminiscent of the equation applied for balancing the training input for text models,
specifically the sampling strategy applied in word2vec~\cite{mikolov2013distributed}.
This resemblance reinforces the importance of our idea that the distribution should not be completely flattened, but rather that differences should be preserved so that the relative ordering of the original data is maintained. 
However, in word2vec the aim is a balance of text tokens using token reduction, and here we are balancing the distribution of samples, not tokens, with a data filtering approach.

\subsection{Dynamic Sampling}

Dynamic sampling randomly selects a different subset of $S_i$ samples from the cluster in each epoch.
In each model pre-training epoch, for each cluster $i$, $S_i$ samples are drawn based on the scaling factor $P_i$ defined as:
\begin{equation}
    P_i = S_i/c_i
    \label{equ:prob}
\end{equation}
This sampling strategy reduces the number of samples seen from large clusters while ensuring that different samples within the cluster can be seen in each epoch.
It increases the number of samples seen from small clusters by upsampling. 
Note that all samples within a cluster share the same probability of being selected.

\section{Experimental Settings}
\label{sec:setup}

We follow OpenCLIP~\cite{gabriel2021openclip}, FLIP~\cite{li2023flip} and CLIPA~\cite{li2023clipa} to pre-train and evaluate our model.

\noindent \textbf{Dataset.}
We pre-train our models on datasets of different sizes. 
Specifically, we use DataComp (Large, 1.28B image-text pairs)~\cite{gadre2023datacomp}, which is filtered with DFN~\cite{fang2023dfn} for the DynamiCS analysis (Sec.\ref{sec:analysis}) and all next experiments. 
We refer to this filtered dataset as DataComp-DFN.
We also conduct DynamiCS on the LAION-400M~\cite{schuhmann2021laion400m} dataset. 
Both datasets are widely used for pre-training VLMs~\cite{cherti2023openclip,li2023flip,li2023reclip,gadre2023datacomp,li2023clipa,fang2023dfn,wei2025hq}. 
Due to expired URLs, we could download about 298 million samples from LAION-400M and 130 million samples from DataComp (out of 192 million candidates).

\noindent  \textbf{Architecture.}
We use ViT-B/16 and ViT-L/16~\cite{dosovitskiy2020ViT} as image encoders and a Transformer model~\cite{vaswani2017Transformer} as the text encoder.
The input image resolutions are $112\times112$ and $224\times224$, and the maximum text length is 32.

\noindent  \textbf{Training and Fine-tuning Setting.}
We pre-train the model on the DataComp and LAION-400M dataset for 1.28 billion and 2.56 billion samples seen, following DFN~\cite{fang2023dfn} and CLIPA~\cite{li2023clipa}.
To accelerate training and conserve computational resources as much as possible,
we first pre-train at a small image resolution (112$\times$112) with a batch size of 28k
which has been shown to be an effective way to speed up CLIP~\cite{li2023reclip,li2023clipa}, and use this as our baseline.
We then fine-tune the model for one additional step with a full-resolution image ($224\times224$) to bridge the distribution gap between training and evaluation.
We set the number of clusters to 50k for both DataComp and LAION-400M (Sec.~\ref{sec:results}). 
We use $\alpha = 0.2$ and set the target number of samples $T$ to 50\% of the dataset size. 
For post-clustering refinement, we use a cosine-similarity threshold of 0.7.

The model was trained on 2 nodes, each with 4 H100 GPUs, with identical settings to ensure that all models were run under consistent conditions.
More details of the training and fine-tuning settings can be found in the Appendix.

\noindent  \textbf{Evaluation}
We adopt the evaluation settings and downstream datasets as CLIP and OpenCLIP~\cite{radford2021clip, cherti2023openclip}.
These datasets cover a broad range of modalities and domains, including natural images, fine-grained classification tasks, and cross-modal retrieval benchmarks to ensure the evaluation is both wide in scope and diverse.
We also evaluate the models on the long-tail concepts dataset \textit{Let It Wag!}~\cite{udandarao2024no}, which consists of 290 long-tail categories curated from multiple VLMs' training datasets and 130K test samples collected from the web (448 images per category). 
For classification, we adopt the 80 prompts introduced in CLIP~\cite{radford2021clip}, which are also used for ImageNet-1K evaluation.

\section{Experimental Analysis of DynamiCS}
\label{sec:analysis}

In this section, we carry out experiments on DataComp to demonstrate the contributions of dynamic sampling and cluster scaling and set key hyperparameters.

\begin{table}[!t]
\caption{\textbf{Zero-shot classification on ImageNet-1K for different $\alpha$ values in Eq.~\ref{equ:shrinkage}}.
All models are pre-trained on \textbf{DataComp} for 106 million samples seen with ViT-B/16 image encoder and $112\times112$ image resolution.
}
\label{tab:alpha}
\footnotesize
\centering
\resizebox{0.5\textwidth}{!}{
\begin{tabular}{l|ccccccc}
\toprule
 & \textbf{0.0} & \textbf{0.2} & \textbf{0.4} & \textbf{0.6} & \textbf{0.8} & \textbf{1.0} & \textbf{2.0} \\
\midrule
\textbf{ImageNet-1K} 
& 38.5 & 39.2 & 38.2 & 36.9 & 36.4 & 33.8 & 19.4 \\
\textbf{\textit{Let It Wag!}}
& 19.5 & 20.2 & 19.6 & 17.5 & 15.6 & 13.4 & 5.1 \\
\bottomrule
\end{tabular}
}

\end{table}

\noindent \textbf{Scaling Factor $\alpha$.}
As shown in Table~\ref{tab:alpha}, the model achieves its best performance when $\alpha = 0.2$.
The result supports the conclusion that, in practice, we need to select a moderate $\alpha$ that balances the short-head and long-tail data. 

Diving into more detail, we see that the ``aim for even'' case, where $\alpha = 0$ and each cluster selects the same number of samples, is outperformed by $\alpha = 0.2$.
This result supports our ``aim for utility'', where the fat head is not completely eliminated and the long-tail is emphasized.
Note that the $\alpha = 0$ still outperforms $\alpha = 1$, which corresponds to the dynamic random sampling approach where all samples have equal probability of being sampled.

When $\alpha = 2$, sampling favors large clusters and suppresses small ones, leading to a highly imbalanced distribution.
Performance drops substantially as $\alpha$ increases from 0.2 to 2, falling from 39.2\% to 19.4\% on ImageNet-1K and from 20.2\% to 5.1\% on \textit{Let It Wag!}.
This drop can be attributed to the loss of representation of long-tailed concepts.

Overall, models trained with $\alpha$ values between 0.0 and 0.8 consistently outperform the random method ($\alpha=1.0$), indicating that DynamiCS is relatively robust to the choice of $\alpha$. 
Based on this robustness, we use $\alpha = 0.2$ for all large-scale experiments in Sec.~\ref{sec:results}, without concern that the experimental results are sensitive to this choice.

\begin{table*}[!t]
\caption{\textbf{Comparison of zero-shot top-1 classification accuracy on ImageNet-1K.}
All models are pre-trained on \textbf{DataComp} for 0.64 billion samples seen with ViT-B/16 image encoder.}
\label{tab:dynamic_datacomp}
\footnotesize
\centering
\resizebox{0.75\textwidth}{!}{
\begin{tabular}{l|c|c|c}
\toprule
\textbf{Models} & \textbf{Samples Seen@Resolution} & \textbf{ImageNet-1K} & \textbf{\textit{Let It Wag!}} \\
\midrule
% \textbf{RECLIP~\cite{li2023flip}}            & 106MB@112 & 81 &  &  \\
\textbf{Random Pruning}                      & \multirow{4}{*}{0.64B@112 + 128M@224} & 64.5  & 35.5  \\
\textbf{Random-Dynamic}  & & 66.2 & 36.2 \\
\textbf{Cluster-Scaling (ours)} & & 68.0  & 43.7   \\
\textbf{DynamiCS (ours)} & & 69.2 & 46.5 \\
\bottomrule
\end{tabular}
}

\end{table*}

\noindent \textbf{Cluster Scaling Helps Improve Long-tail Concepts Performance.}
As shown in Table~\ref{tab:dynamic_datacomp}, cluster scaling substantially improves pre-training under both random and dynamic sampling. 
It substantially increases performance by 3.5\% and 3.0\% on ImageNet-1K, and by 8.2\% and 10.3\% on \textit{Let It Wag!}, respectively.
The improvement can be attributed to DynamiCS enhancing the performance of long-tail concepts while maintaining that of head concepts.

\noindent \textbf{Dynamic Sampling Increases Training Data Diversity.}
Then, we show the influence of dynamic sampling on CLIP pre-training. 
The results in Table~\ref{tab:dynamic_datacomp} show the improvement that dynamic sampling delivers compared to standard, static sampling.
DynamiCS outperforms the Random pruning method by 4.7\% and 11.0\% on the ImageNet-1K and \textit{Let It Wag!} datasets.
Here, random pruning means that it randomly selects a fixed 50\% subset of the dataset.
Random-Dynamic outperforms random pruning by 1.7\% and 0.7\% on the ImageNet-1K and \textit{Let It Wag!} datasets.
Here, Random-Dynamic means that each sample has a 50\% probability of being selected during training.
And DynamiCS outperforms Cluster-Scaling by 1.2\% and 2.8\% on the ImageNet-1K and \textit{Let It Wag!} datasets.
The gains of dynamic sampling can be attributed to an improvement in the diversity of the data used for training, compared with static sampling.

\section{Comparative Experimental Results}
\label{sec:results}

In this section, we present an evaluation of DynamiCS trained on LAION-400M~\cite{schuhmann2021laion400m} and DataComp~\cite{gadre2023datacomp,fang2023dfn} on the classic zero-shot ImageNet-1K task, on a long-tail task, and on a range of other datasets conventionally used in the literature to evaluate CLIP performance. 

\begin{table*}[t!]
\caption{\textbf{Random sampling vs.\ DynamiCS.} Results on ImageNet-1K and \textit{Let It Wag!} under same training settings on LAION-400M and DataComp-DFN. 
% All models use the ViT-B/16 image encoder.
}
\label{tab:imagenet_random_vs_dynamics}
\footnotesize
\centering
\resizebox{\textwidth}{!}{
\begin{tabular}{l|c|c|c|c|c}
\toprule
\textbf{Models} & \textbf{\makecell[c]{Dataset \\ (Data Size)}} & \textbf{Samples Seen@Resolution} &
\textbf{ImageNet-1K} & \textbf{\textit{Let It Wag!}} & \textbf{GPU hours} \\
\midrule

\textbf{Random} & \multirow{2}{*}{\makecell[c]{LAION-400M \\ (298M)}} & 1.28B@112 + 128M@224 & 59.8 & 31.9 & 151 \\
\textbf{DynamiCS (Ours)} &  & 1.28B@112 + 128M@224 & 65.0 & 42.1 & 163 \\
\midrule

\textbf{Random} & \multirow{2}{*}{\makecell[c]{DataComp-DFN \\ (130M)}} & 0.64B@112 + 128M@224 & 64.5 & 35.5 & 90 \\
\textbf{DynamiCS (Ours)} &  & 0.64B@112 + 128M@224 & 69.2 & 46.5 & 95 \\
\bottomrule
\end{tabular}
}

\end{table*}

\subsection{DynamiCS \textit{vs.} Random Sampling}

As shown in Table~\ref{tab:imagenet_random_vs_dynamics}, DynamiCS outperforms random pruning by 5.2\% and 4.7\% on ImageNet-1K when pre-training on LAION-400M and DataComp, respectively. 
The improvements on the long-tail benchmark are larger (10.2\% and 11.0\%), 
indicating that DynamiCS improves coverage of long-tail concepts while also delivering clear improvements on ImageNet-1K.

\begin{table*}[t!]
\caption{\textbf{Comparison of zero-shot top-1 classification on ImageNet-1K.} 
Our model was pre-trained on the LAION-400M dataset and a subset of \textbf{DataComp-Large} that was filtered by DFN-2B.
FLIP is pre-trained with 75\% image masking, resulting in the same number of image tokens as a 112$\times$112 image size.
CLIPA is pre-trained by syntax masking with 16 text tokens.
The symbol $*$ indicates results we reproduced under the same low-resolution and training setting to enable fairer comparison wherever possible.
The symbol $\approx$ indicates estimated values, because GPU hours are not reported in their paper.
All models use the ViT-B/16 image encoder.
}
\label{tab:imagenet_other_methods}
\footnotesize
\centering
\resizebox{\textwidth}{!}{
\begin{tabular}{l|c|c|c|c|c}
\toprule
\textbf{Models} & \textbf{\makecell[c]{Dataset \\ (Data Size)}} & \textbf{Samples Seen@Resolution} &
\textbf{ImageNet-1K} & \textbf{\textit{Let It Wag!}} & \textbf{GPU hours} \\
\midrule
\textbf{OpenCLIP~\cite{cherti2023openclip}} & \multirow{2}{*}{\makecell[c]{LAION-400M \\ (400M)}}  & 2.56B@224 & 64.2 & --- & 2140 \\
\textbf{FLIP~\cite{li2023flip}} &  & 2.56B@224 + 128M@224 & 60.9 & --- & --- \\ \cline{0-0} \cline{2-6}
\textbf{CLIPA*~\cite{li2023clipa}} &  \multirow{4}{*}{\makecell[c]{LAION-400M \\ (298M)}} & 2.56B@112 + 128M@224 & 63.2 & 36.4 & 269 \\
\textbf{RECLIP*~\cite{li2023reclip}} &  & 2.56B@112 + 128M@224 & 62.9 & 36.0 & 280 \\ \cline{0-0} \cline{3-6}
\textbf{DynamiCS (Ours)} &  & 1.28B@112 + 128M@224 & \textbf{65.0} & 42.1 & 163 \\
\textbf{DynamiCS (Ours)} &  & 2.56B@112 + 128M@224 & \textbf{67.5} & 45.5 & 299 \\
\midrule
\textbf{DataComp~\cite{gadre2023datacomp}} & \multirow{2}{*}{\makecell[c]{DataComp \\ (1.28B)}} & 1.28B@224 & 63.1 & 33.7 & $\approx1070$ \\
\textbf{Captioning~\cite{nguyen2023improving}} &  & 2.56B@224 & 59.8 & --- & $\approx2140$ \\ 
\hline
\multirow{2}{*}{\textbf{WhatIf}~\cite{li2025whatif}} & \multirow{1}{*}{\makecell[c]{Recap-DataComp-1B \\ (1.4B)}} & \multirow{2}{*}{2.56B@112 + 128M@224} & \multirow{2}{*}{69.2} & \multirow{2}{*}{---} & \multirow{2}{*}{---} \\ \\
% \textbf{OpenVision} &  & 12.8B@160 + 1.024B@224 + 256M@336 & 73.9 & --- & --- \\ 
\hline
\textbf{DFN~\cite{fang2023dfn}} &  \multirow{5}{*}{\makecell[c]{DataComp-DFN \\ (130M)}} & 1.28B@224 & 67.8 & --- & $\approx1070$ \\
\textbf{HQ-CLIP~\cite{wei2025hq}} & & 3.20B@224 & 70.6 & 38.2 & $\approx2675$ \\
\textbf{DFN*~\cite{fang2023dfn}} &  & 1.28B@112 + 128M@224 & 68.7 & 42.4 & 151 \\ \cline{0-0} \cline{3-6}
\textbf{DynamiCS (Ours)} &  & 0.64B@112 + 128M@224 & 69.2 & 46.5 & 95 \\
\textbf{DynamiCS (Ours)} &  & 1.28B@112 + 128M@224 & \textbf{71.3} & \textbf{50.2} & 163 \\
\bottomrule
\end{tabular}
}

\end{table*}

\subsection{Zero-shot Classification on ImageNet-1K}

\noindent \textbf{Cost-saving Baselines on LAION-400M:}
DynamiCS outperforms cost-saving baselines, RECLIP~\cite{li2023reclip}, FLIP~\cite{li2023flip}, CLIPA~\cite{li2023clipa} as demonstrated in Table~\ref{tab:imagenet_other_methods}, with around 50\% of the samples seen and 60\% GPU hours used by baselines.
Specifically, DynamiCS-1.28B outperforms CLIPA by 1.8\%, RECLIP by 2.1\% and FLIP by 4.1\% on ImageNet-1K while using only 50\% of the total samples seen, primarily due to improved performance on long-tail concepts (See Appendix).
Notably, our dynamic sampling plays a key role in reducing training cost while preserving model quality.

\noindent \textbf{Dual-purpose Filtering-based Baselines on DataComp:}
DynamiCS also outperforms different dual-purpose filtering-based baselines.
We first reproduce the result for DFN~\cite{fang2023dfn} on the filtered DataComp-Large dataset, \textbf{DFN*} in Table~\ref{tab:imagenet_other_methods}.
We then apply DynamiCS on the same filtered dataset. 
DynamiCS-0.64B improves over DFN* by 0.5\% on ImageNet-1K while using only 50\% of the training samples and 62\% of the GPU hours. 
DynamiCS-1.28B also outperforms HQ-CLIP by 0.7\% with about 40\% of the samples and about 6\% of the GPU hours. 
In addition, DynamiCS-0.64B surpasses the Captioning baseline by 9.4\% on ImageNet-1K.
DynamiCS achieves comparable or better ImageNet-1K performance than WhatIf~\cite{li2025whatif} (recaptioning on DataComp-1B), while requiring substantially less computing resources.

\subsection{Zero-shot Classification on Long-tail Dataset}
\label{sec:results_long_tail}

To further investigate whether upsampling long-tail concepts can improve the performance of low-frequency classes.  
We further evaluate DynamiCS on the long tail dataset, which is \textit{Let It Wag!}~\cite{udandarao2024no}, as shown in Table~\ref{tab:imagenet_other_methods}
demonstrates that DynamiCS can substantially enhance the long-tail performance, thereby contributing to the model's overall performance.

DynamiCS-1.28B pre-trained on LAION-400M outperforms the baseline model RECLIP by 6.1\% and CLIPA by 5.7\% on the \textit{Let It Wag!} dataset, while maintaining comparable performance on ImageNet and using only about 50\% of the total samples seen and 60\% GPU hours.
Furthermore, DynamiCS-0.64B pre-trained on DataComp-DFN substantially outperforms the DataComp (Image-based $\cap$ CLIP score) filtering methods that are pre-trained on the DataComp dataset by 12.8\%.
DynamiCS-0.64B also outperforms HQ-CLIP by 8.3\% on the long-tail dataset with only 20\% samples seen and 3.6\% GPU hours.
Overall, DynamiCS boosts long-tail performance at far lower training cost, outperforming large-scale models and baselines by a wide margin.

\subsection{Data Scaling}

\begin{table*}[t!]
\caption{\textbf{Data scaling results for DynamiCS.} 
Zero-shot top-1 accuracy on ImageNet-1K and \textit{Let It Wag!} when scaling the number of training samples from 0.64 billion to 2.56 billion samples seen on LAION-400M and DataComp-DFN. 
}
\label{tab:imagenet_dynamics_scaling}
\footnotesize
\centering
\resizebox{\textwidth}{!}{
\begin{tabular}{l|c|c|c|c|c}
\toprule
\textbf{Models} & \textbf{\makecell[c]{Dataset \\ (Data Size)}}  & \textbf{Samples Seen@Resolution} &
\textbf{ImageNet-1K} & \textbf{\textit{Let It Wag!}} & \textbf{GPU hours} \\
\midrule
\multirow{6}{*}{\textbf{DynamiCS (Ours)}}
& \multirow{3}{*}{\makecell[c]{LAION-400M \\ (298M)}}
& 0.64B@112 + 128M@224 & 61.5 & 37.3 & 95 \\
&  & 1.28B@112 + 128M@224 & 65.0 & 42.1 & 163 \\
&  & 2.56B@112 + 128M@224 & 67.5 & 45.5 & 299 \\
\cline{2-6}
& \multirow{3}{*}{\makecell[c]{DataComp-DFN \\ (130M)}}
& 0.64B@112 + 128M@224 & 69.2 & 46.5 & 95 \\
&  & 1.28B@112 + 128M@224 & 71.3 & 50.2 & 163 \\
&  & 2.56B@112 + 128M@224 & \textbf{72.6} & \textbf{52.0} & 299 \\
\bottomrule
\end{tabular}
}

\end{table*}

DynamiCS has shown a substantial improvement over random pruning, cost-saving baselines, and dual-purpose baselines on both ImageNet-1K and \textit{Let It Wag!}, while using about 60\% of the GPU hours.
We further scale DynamiCS from 0.64B to 2.56B samples seen on LAION-400M and DataComp. 
As shown in Table~\ref{tab:imagenet_dynamics_scaling},
on LAION-400M, ImageNet-1K accuracy increases from 61.5\% to 67.5\%, and \textit{Let It Wag!} accuracy increases from 37.3\% to 45.5\%. 
On DataComp, ImageNet-1K accuracy improves from 69.2\% to 72.6\%, and \textit{Let It Wag!} improves from 46.5\% to 52.0\%. 
Overall, the scaling experiments show a clear and consistent improvement as the number of training samples increases.

\begin{table*}[!t]
\caption{\textbf{Zero-shot top-1 classification on ImageNet-1K and \textit{Let It Wag!} with the full training CLIP.} 
All models use the ViT-B/16 image encoder.
}
\label{tab:full_clip_baselines}
\footnotesize
\centering
\resizebox{\textwidth}{!}{
\begin{tabular}{l|c|c|c|c|c|c}
\toprule
\textbf{Models} & \textbf{\makecell[c]{Dataset \\ (Data Size)}}  & \textbf{Samples Seen@Resolution} & \textbf{Tokens} &
\textbf{ImageNet-1K} & \textbf{\textit{Let It Wag!}} & \textbf{GPU hours} \\
\midrule
\textbf{OpenAI-WIT~\cite{radford2021clip}} & \multirow{2}{*}{\makecell[c]{---\\ (400M)}} & 12.8B@224 & 274 & 68.3 & 37.9 & 10700 \\
\textbf{MetaCLIP-400M~\cite{xu2024metaclip}} &  & 12.8B@224 & 274 & 70.8 & 46.5 & $\approx10700$ \\ \hline
\textbf{OpenCLIP~\cite{cherti2023openclip} }& \multirow{2}{*}{\makecell[c]{LAION-400M \\ (400M)}} & 12.8B@224 & 274 & 67.1 & 39.1 & 10736 \\
\textbf{LaCLIP~\cite{fan2023laclip}} &  & 12.8B@224 & 274 & 69.4 & 48.4 & $\approx10700$ \\  \hline
 \multirow{2}{*}{\textbf{DynamiCS (Ours)}} &  \multirow{1}{*}{\makecell[c]{LAION-400M \\ (298M)}} &  \multirow{2}{*}{2.56B@112 + 128M@224} &  \multirow{2}{*}{~~81} &  \multirow{2}{*}{67.5} &  \multirow{2}{*}{45.5} &  \multirow{2}{*}{299} \\ \\ 
\midrule
\multirow{2}{*}{\textbf{OpenVision~\cite{li2025openvision}}} & \multirow{1}{*}{\makecell[c]{Recap-DataComp-1B \\ (1.4B)}}  & \multirow{2}{*}{{\makecell[c]{12.8B@160 + 1.024B@224 \\ + 256M@336}} } & \multirow{2}{*}{180} & \multirow{2}{*}{73.9} & \multirow{2}{*}{---} & \multirow{2}{*}{---} \\  \\
\hline
\multirow{2}{*}{\textbf{DynamiCS (Ours)}} & \multirow{2}{*}{\makecell[c]{DataComp-DFN \\ (130M)}} 
  & 1.28B@112 + 128M@224 & ~~81 & 71.3 & 50.2 & 163 \\
& & 2.56B@112 + 128M@224 & ~~81 & 72.6 & 52.0 & 299 \\
\bottomrule
\end{tabular}
}

\end{table*}

\subsection{Comparison with Full-scale Training CLIP Baselines} 

As shown in Table~\ref{tab:full_clip_baselines}, we compare DynamiCS with the full-scale training CLIP baselines, which are pre-trained with $224\times224$ image resolution and 12.8 billion samples seen.
Our DynamiCS-2.56B has the best performance on both ImageNet-1K and \textit{Let It Wag!} datasets.
Surprisingly, DynamiCS pre-trained on DataComp with 2.56B samples seen, with only about 3\% of the training cost, outperforms OpenAI-WIT CLIP, OpenCLIP, and MetaCLIP
by 4.3\%, 5.5\%, 1.8\% on ImageNet-1K and by 14.1\%, 12.9\%, 5.5\% on the \textit{Let It Wag!} dataset.
Moreover, DynamiCS-1.28B pre-trained on DataComp with 1.28B samples already surpasses full-training baselines using only 163 GPU hours.

\subsection{Zero-shot Robustness}

We evaluate DynamiCS on 6 robustness datasets following the evaluation of CLIP~\cite{radford2021clip} and OpenCLIP~\cite{cherti2023openclip} as shown in Table~\ref{tab:imagenet_comparison}.
On LAION-400M, DynamiCS-1.28B outperforms CLIPA by 1.1\% and RECLIP by 1.5\%, while using only 50\% of the samples seen by RECLIP. 
It also improves over the random baseline by 4.6\%. 
DynamiCS-2.56B remains below full-training CLIP, with a gap of 0.4\%.

On DataComp, DynamiCS performs well compared with CLIP-score-based filtering methods. 
DynamiCS-0.64B already outperforms DataComp by 4.9\% while using only 50\% of the samples and 8\% of the GPU hours.
DynamiCS-1.28B improves over the DFN baseline by 2.2\% and over HQ-CLIP by 1.0\%. 
Both DynamiCS-1.28B and DynamiCS-2.56B also outperform fully trained OpenAI-WIT, MetaCLIP, and OpenCLIP. 
In particular, DynamiCS-2.56B exceeds full training on OpenAI-WIT by 3.2\%, MetaCLIP by 1.6\%, and OpenCLIP by 5.0\%, while using only 3\% of the GPU hours.
Overall, DynamiCS achieves comparable robustness results while using substantially fewer GPU hours, demonstrating strong robustness and improved data efficiency.

\begin{table*}[!t]
\caption{\textbf{Zero-shot robustness evaluation} of DynamiCS and other
methods on the different robustness datasets.
}
\label{tab:imagenet_comparison}
\footnotesize
\centering
\resizebox{\textwidth}{!}{
\begin{tabular}{l|c|c|cccccc|c|c}
\toprule
\textbf{Models} & \textbf{\makecell[c]{Dataset \\ (Data Size)}} & \textbf{Samples Seen@Resolution} &
\textbf{IN-A} & \textbf{IN-O} & \textbf{IN-R} & \textbf{IN-S} & \textbf{IN-V2} & \textbf{ON} &
\textbf{Avg.} & \textbf{GPU hours}  \\
\midrule

\textbf{OpenAI-WIT~\cite{radford2021clip}} 
& \multirow{2}{*}{\makecell[c]{---\\ (400M)}} 
& 12.8B@224  
& 50.0 & 42.3 & 77.7 & 48.2 & 55.3 & 61.9 & 55.9 & 10700 \\

\textbf{MetaCLIP-400M~\cite{xu2024metaclip}} 
&  
& 12.8B@224  
& -- & -- & -- & -- & -- & -- & 57.5 & $\approx10700$ \\

\midrule

\multirow{2}{*}{\textbf{OpenCLIP~\cite{cherti2023openclip}}}
&  \multirow{2}{*}{\makecell[c]{LAION-400M \\ (400M)}} 
&  \multirow{2}{*}{12.8B@224}  
&  \multirow{2}{*}{33.2} &  \multirow{2}{*}{50.8} &  \multirow{2}{*}{77.9} &  \multirow{2}{*}{52.4} &   \multirow{2}{*}{50.8} &  \multirow{2}{*}{59.6} &  \multirow{2}{*}{54.1} &  \multirow{2}{*}{10736} \\   \\
\midrule

\textbf{RECLIP*~\cite{li2023reclip}} 
&  \multirow{5}{*}{\makecell[c]{LAION-400M \\ (298M)}} 
& 2.56B@112 + 128M@224 
& 26.1 & 53.5 & 72.8 & 48.1 & 55.2 & 47.2 & 50.5 & 280 \\

\textbf{CLIPA*~\cite{li2023clipa}} 
&  
& 2.56B@112 + 128M@224 
& 26.7 & 54.1 & 73.3 & 48.6 & 55.5 & 47.2 & 50.9 & 269 \\

\cmidrule(lr){3-11}
\textbf{Random} 
& 
& 1.28B@112 + 128M@224 
& 21.6 & 53.7 & 69.3 & 44.9 & 52.0 & 43.1 & 47.4 & 151 \\ 

\textbf{DynamiCS (Ours)} 
& 
& 1.28B@112 + 128M@224 
& 28.9 & 56.1 & 73.2 & 49.2 & 56.4 & 47.9 & 52.0 & 163 \\

\cmidrule(lr){3-11}
\textbf{DynamiCS (Ours)} 
& 
& 2.56B@112 + 128M@224 
& 30.9 & 53.5 & 76.3 & 51.7 & 59.2 & 50.7 & 53.7 & 299 \\

\midrule

\textbf{DataComp~\cite{gadre2023datacomp}} 
& \makecell[c]{DataComp-Large \\ (1.28B)} 
& 1.28B@224  
& 25.5 & 49.6 & 71.8 & 49.8 & 55.1 & 53.1 & 50.8 & $\approx1070$ \\

\midrule

\textbf{DFN~\cite{fang2023dfn}} 
& \multirow{7}{*}{\makecell[c]{DataComp-DFN \\ (130M)}}  
& 1.28B@224  
& -- & -- & -- & -- & -- & -- & 54.0  & $\approx1070$ \\

\textbf{HQ-CLIP~\cite{wei2025hq}} 
&  
& 3.20B@224  
& 39.1 & 43.0 & 80.1 & 57.3  & 63.1 & 60.6 & 57.2 & $\approx2675$ \\ 

\textbf{DFN*~\cite{fang2023dfn}} 
&  
& 1.28B@112 + 128M@224 
& 31.4 & 59.7 & 74.5 & 53.9 & 60.9 & 55.5 & 56.0 & 151 \\

\cmidrule(lr){3-11}
\textbf{Random} 
&  
& 0.64B@112 + 128M@224 
& 23.0 & 59.5 & 70.3 & 49.3 & 56.4 & 51.4 & 51.7 & 90 \\

\textbf{DynamiCS (Ours)} 
&  
& 0.64B@112 + 128M@224 
& 30.4 & 62.3 & 73.2 & 53.2 & 60.7 & 54.6 & 55.7 & 95 \\  

\cmidrule(lr){3-11}
\textbf{DynamiCS (Ours)} 
&  
& 1.28B@112 + 128M@224 
& 35.7 & 58.4 & 76.6 & 55.9 & 64.0 & 58.4 & 58.2 & 163 \\

\textbf{DynamiCS (Ours)} 
&  
& 2.56B@112 + 128M@224 
& 37.2 & 57.6 & 79.0 & 57.4 & 64.1 & 59.3 & 59.1 & 299 \\
\bottomrule
\end{tabular}
}

\end{table*}

\begin{table*}[t!]
\caption{\textbf{Zero-shot image-text retrieval evaluation} of DynamiCS and other methods on the COCO~\cite{lin2014MSCOCO} and Flickr30k~\cite{young2014Flickr30k}. 
Report the average recall@1 of the image-to-text and text-to-image retrieval results.
% All models use the ViT-B/16 image encoder.
}
\label{tab:retrieval}
\footnotesize
\centering
\resizebox{0.9\textwidth}{!}{
\begin{tabular}{l|c|c|cc|c}
\toprule
\multirow{2}{*}{\textbf{Models}} 
& \multirow{2}{*}{\textbf{\makecell[c]{Dataset \\ (Data Size)}}} 
& \multirow{2}{*}{\textbf{Samples Seen@Resolution}}
& \multirow{2}{*}{\textbf{COCO}} &  \multirow{2}{*}{\textbf{Flickr30k}}
& \multirow{2}{*}{\makecell{\textbf{GPU hours}}} \\ \\
\midrule

\textbf{OpenAI-WIT~\cite{radford2021clip}} 
& \multirow{2}{*}{\makecell[c]{---\\ (400M)}} 
& 12.8B@224  
& 42.8 & 72.2 & 10700 \\

\textbf{MetaCLIP-400M~\cite{xu2024metaclip}} 
&  
& 12.8B@224  
& 48.2 & 76.7 & $\approx10700$ \\

\midrule

\textbf{OpenCLIP~\cite{gabriel2021openclip}} 
& \multirow{2}{*}{\makecell[c]{LAION-400M \\ (400M)}} 
& 12.8B@224  
& 46.9 & 74.6 & 10736\\  

\textbf{FLIP~\cite{li2023flip}} 
&  
& 2.56B@224 + 128M@224  
& -- & -- & ---  \\

\midrule

\textbf{RECLIP*~\cite{li2023reclip}} 
& \multirow{3}{*}{\makecell[c]{LAION-400M \\ (298M)}} 
& 2.56B@112 + 128M@224  
& 44.4 & 72.4 & 280 \\

\textbf{CLIPA*~\cite{li2023clipa}} 
&  
& 2.56B@112 + 128M@224  
& 44.4 & 71.7 & 269 \\

\cmidrule(lr){3-6}
\textbf{Random} 
& 
& 1.28B@112 + 128M@224  
& 42.7 & 69.4 & 151 \\

\textbf{DynamiCS (Ours)} 
& 
& 1.28B@112 + 128M@224  
& 45.1 & 72.7 & 163 \\

\cmidrule(lr){3-6}
\textbf{DynamiCS (Ours)} 
& 
& 2.56B@112 + 128M@224  
& 46.5 & 73.9 & 299 \\

\midrule

\textbf{DataComp~\cite{gadre2023datacomp}} 
& \makecell[c]{DataComp-Large \\ (1.28B)} 
& 1.28B@224  
& 39.8 & 63.6 & $\approx1070$ \\

\midrule

\textbf{HQ-CLIP~\cite{wei2025hq}} 
& \multirow{6}{*}{\makecell[c]{DataComp-DFN \\ (130M)}} 
& 3.20B@224  
& 52.2 & 77.9 & $\approx2675$ \\ 

\textbf{DFN*~\cite{fang2023dfn}} 
&  
& 1.28B@112 + 128M@224  
& 44.3 & 68.1 & 151 \\

\cmidrule(lr){3-6}
\textbf{Random} 
&  
& 0.64B@112 + 128M@224  
& 39.6 & 62.3 & 90 \\

\textbf{DynamiCS (Ours)} 
&  
& 0.64B@112 + 128M@224  
& 42.5 & 66.6 & 95 \\  

\cmidrule(lr){3-6}
\textbf{DynamiCS (Ours)} 
&  
& 1.28B@112 + 128M@224  
& 44.2 & 69.3 & 163 \\

\textbf{DynamiCS (Ours)} 
&  
& 2.56B@112 + 128M@224  
& 46.2 & 71.3 & 299 \\
\bottomrule
\end{tabular}
}

\end{table*}

\subsection{Image-Text Retrieval}

Table~\ref{tab:retrieval} reports zero-shot retrieval results on COCO~\cite{lin2014MSCOCO} and Flickr30k~\cite{young2014Flickr30k}.
Overall, DynamiCS consistently outperforms RECLIP, CLIPA, and the Random baseline on both LAION-400M and DataComp.
On LAION-400M, DynamiCS-2.56B exceeds fully trained OpenAI-WIT, but remains below OpenCLIP and MetaCLIP. 
In contrast to the zero-shot classification results, pre-training on DataComp does not provide a clear advantage over LAION-400M for image–text retrieval.
HQ-CLIP outperforms DynamiCS, likely because its generated captions are longer and more detailed, which better match the needs of retrieval tasks.
DynamiCS can also be applied to HQ-CLIP’s synthetic data to further improve retrieval performance, since HQ-CLIP does not explicitly balance the data.

\begin{table*}[t!]
\caption{\textbf{Zero-shot top-1 classification accuracy across 22 benchmark classification datasets.}
}
\label{tab:dataset_performance}
\footnotesize
\centering
\resizebox{\textwidth}{!}{
\begin{tabular}{l|c|c|*{22}{c}|c}
\toprule
\textbf{Models} & \textbf{\makecell[c]{Dataset \\ (Data Size)}} & \makecell[c]{\textbf{Samples Seen } \\ \textbf{@Resolution}}
& \rotatebox{90}{Food-101}
& \rotatebox{90}{CIFAR-10}
& \rotatebox{90}{CIFAR-100}
& \rotatebox{90}{CUB200}
& \rotatebox{90}{SUN397}
& \rotatebox{90}{Cars}
& \rotatebox{90}{Aircraft}
& \rotatebox{90}{VOC2007}
& \rotatebox{90}{DTD}
& \rotatebox{90}{OxfordPets}
& \rotatebox{90}{Caltech-101}
& \rotatebox{90}{Flowers102}
& \rotatebox{90}{MNIST}
& \rotatebox{90}{STL10}
& \rotatebox{90}{EuroSAT}
& \rotatebox{90}{Resisc45}
& \rotatebox{90}{GTSRB}
& \rotatebox{90}{KITTI}
& \rotatebox{90}{Country211}
& \rotatebox{90}{PCAM}
& \rotatebox{90}{CLEVR}
& \rotatebox{90}{SST2}
& \rotatebox{90}{\textbf{GPU hours}} \\
\midrule

\textbf{OpenAI-WIT~\cite{radford2021clip}} & \multirow{2}{*}{\makecell[c]{---\\ (400M)}} & 12.8B@224
& 88.7 & 90.8 & 67.0 & --- & 64.8 & 58.2 & 24.2 & 78.3 & 45.0 & 88.9 & 89.0 & --- & 51.4 & 98.3 & 55.9 & 60.7 & 43.4 & 26.4 & 22.8 & 50.7 & 21.2 & 50.7
& 10700 \\

\textbf{MetaCLIP-400M~\cite{xu2024metaclip}} &  & 12.8B@224
& 87.3 & 90.1 & 66.6 & --- & 66.8 & 74.2 & 28.4 & 72.2 & 55.9 & 90.4 & 93.4 & 72.3 & 47.9 & 97.2 & 55.7 & 66.2 & 43.8 & 24.2 & 22.6 & 62.0 & 30.1 & 62.0
& $\approx10700$ \\
\midrule

\textbf{OpenCLIP~\cite{cherti2023openclip}} & \multirow{4}{*}{\makecell[c]{LAION-400M \\ (298M)}} & 12.8B@224
& 86.1 & 91.7 & 71.2 & --- & 69.6 & --- & 17.7 & 76.8 & 51.3 & 89.2 & 91.3 & --- & 66.2 & --- & 50.2 & 58.5 & 43.5 & 18.1 & 18.1 & 59.6 & 28.7 & 54.4
& 10736 \\

\textbf{RECLIP*~\cite{li2023reclip}} &  & 2.56B@112 + 128M@224
& 81.3 & 92.1 & 71.8 & 55.1 & 66.0 & 80.7 & 13.0 & 75.3 & 49.4 & 85.3 & 82.1 & 64.6 & 65.7 & 96.3 & 45.5 & 55.5 & 33.5 & 21.7 & 14.4 & 48.9 & 17.7 & 49.8
& 280 \\

\textbf{CLIPA*~\cite{li2023clipa}} &  & 2.56B@112 + 128M@224
& 81.5 & 91.8 & 70.3 & 57.2 & 65.8 & 81.1 & 14.3 & 76.6 & 46.8 & 85.9 & 82.8 & 61.7 & 41.8 & 96.3 & 55.9 & 55.5 & 38.7 & 21.2 & 14.5 & 50.7 & 18.8 & 53.0
& 269 \\
\cmidrule(lr){1-1}\cmidrule(lr){3-26}

\textbf{DynamiCS (Ours)} & & 2.56B@112 + 128M@224
& 84.0 & 92.9 & 75.0 & 68.9 & 67.5 & 75.7 & 16.8 & 76.8 & 52.9 & 87.7 & 83.7 & 70.4 & 53.1 & 96.3 & 42.9 & 59.1 & 44.1 & 15.6 & 16.2 & 48.2 & 14.0 & 52.9
& 299 \\

\midrule

\textbf{DataComp~\cite{gadre2023datacomp}} & \makecell[c]{DataComp \\ (1.28B)} & 1.28B@224
& 83.1 & 93.8 & 75.4 & 47.0 & 64.3 & 77.2 & 10.0 & 80.9 & 46.9 & 83.5 & 89.7 & 64.0 & 54.0 & 95.8 & 50.1 & 52.7 & 43.4 & 40.1 & 14.3 & 49.7 & 23.1 & 52.9
& $\approx1070$ \\
\midrule

\textbf{HQ-CLIP~\cite{wei2025hq}} & \multirow{6}{*}{\makecell[c]{DataComp-DFN \\ (130M)}} & 3.20B@224
& 87.8 & 96.2 & 81.0 & --- & 69.7 & 85.3 & 11.3 & 78.8 & 51.5 & 89.5 & 93.1 & 69.0 & 77.7 & 98.1 & 47.6 & 60.6 & 54.4 & 43.0 & 15.9 & 47.5 & 27.5 & 51.7
& $\approx2675$ \\

\textbf{DFN*~\cite{fang2023dfn}} &  & 1.28B@112 + 128M@224
& 85.8 & 95.1 & 79.4 & 58.1 & 66.1 & 86.6 & 14.2 & 76.7 & 46.3 & 89.9 & 84.8 & 73.2 & 64.2 & 96.8 & 46.2 & 48.6 & 42.2 & 35.7 & 12.3 & 52.6 & 17.7 & 49.5
& 151 \\
\cmidrule(lr){1-1}\cmidrule(lr){3-26}

\textbf{Random} &  & 0.64B@112 + 128M@224
& 82.7 & 93.7 & 77.3 & 50.7 & 62.9 & 83.7 & 10.1 & 76.9 & 40.9 & 86.7 & 84.0 & 70.6 & 53.3 & 95.7 & 47.3 & 42.7 & 22.4 & 30.8 & 10.8 & 57.7 & 14.9 & 50.4
& 90 \\

\textbf{DynamiCS (Ours)} &  & 0.64B@112 + 128M@224
& 84.6 & 94.0 & 79.5 & 65.2 & 65.4 & 74.7 & 14.2 & 76.4 & 42.0 & 88.4 & 84.5 & 75.9 & 41.2 & 96.0 & 45.8 & 44.8 & 25.5 & 21.1 & 12.1 & 61.6 & 14.7 & 46.6
& 95 \\
\cmidrule(lr){3-26}

\textbf{DynamiCS (Ours)} &  & 1.28B@112 + 128M@224
& 86.9 & 95.2 & 81.3 & 70.8 & 66.7 & 78.3 & 16.5 & 76.5 & 45.4 & 89.9 & 84.3 & 83.5 & 53.2 & 96.7 & 48.9 & 55.5 & 31.1 & 11.7 & 13.5 & 42.1 & 22.2 & 50.1
& 163 \\

\textbf{DynamiCS (Ours)} &  & 2.56B@112 + 128M@224
& 86.7 & 96.2 & 81.8 & 71.3 & 68.1 & 80.8 & 19.2 & 74.6 & 49.2 & 90.7 & 84.4 & 79.0 & 66.8 & 97.3 & 44.2 & 56.7 & 41.5 & 31.5 & 14.4 & 56.5 & 27.2 & 48.9
& 299 \\

\bottomrule
\end{tabular}
}

\end{table*}

\subsection{Zero-shot Classification on Other Datasets}

DynamiCS consistently outperforms random pruning and achieves performance comparable to RECLIP and DFN, as shown in Table~\ref{tab:dataset_performance}. 
Interestingly, compared with these baselines, DynamiCS shows a clear advantage on fine-grained recognition tasks such as CUB-200-2011~\cite{WahCUB_200_2011}, which contains 200 bird species primarily from North America. 
On LAION-400M, DynamiCS-2.56B outperforms RECLIP by 13.8\% and CLIPA by 11.7\%.
On DataComp, DynamiCS-0.64B exceeds the random baseline by 14.5\%, and DynamiCS-1.28B surpasses DFN* by 12.7\%.

On the other fine-grained Flowers102 dataset~\cite{Nilsback08oxford_flower} with 102 flower categories.
On LAION-400M, DynamiCS again improves effectiveness substantially, outperforming RECLIP by 5.8\% and CLIPA by 8.7\%. 
On DataComp, DynamiCS-0.64B outperforms random by 5.3\% and DynamiCS-1.28B outperforms DFN* by 10.3\%.
DynamiCS-1.28B also improves substantially over the results of models filtered and pre-trained on the DataComp dataset, by 19.5\%. 
We leave further exploration to future work.

\subsection{Larger Vision Transformer}

To demonstrate scalability, we scale the model from ViT-B/16 to ViT-L/16, which has about $2.9\times$ more parameters.
As shown in Table~\ref{tab:imagenet_vit_l16}, DynamiCS maintains strong performance and outperforms CLIPA by 1.5\% on ImageNet, while using only 50\% of the samples seen by CLIPA.

\begin{table}[t!]
\caption{\textbf{Zero-shot top-1 classification accuracy with ViT-L/16 encoder on ImageNet-1K.} 
We pre-train the model at an image resolution of $112\times112$ with 16 text tokens (using syntax masking) for 1.28B samples seen, and then fine-tune it in small steps, reducing the total samples seen to 50\% of that of CLIPA.
}
\label{tab:imagenet_vit_l16}
\footnotesize
\centering
\resizebox{0.9\linewidth}{!}{%
\begin{tabular}{l|c|c|c|c|c}
\toprule
\multirow{2}{*}{\textbf{Models}} & 
\multirow{2}{*}{\makecell[c]{\textbf{Dataset} \\ \textbf{(Data Size)}}} &
\multirow{2}{*}{\makecell[c]{\textbf{Samples Seen } \\ \textbf{@Resolution}}} &
\multicolumn{2}{c}{\textbf{ViT-L/16}} & \textbf{GPU hours} \\ 
    & & & \textbf{ImageNet-1K} &  \textbf{\textit{Let It Wag!}} & \\
\midrule

\textbf{CLIPA~\cite{li2023clipa}} 
& \makecell[c]{LAION-400M \\ (400M)} 
& 2.56B@112 + 128M@224 
& 68.8 & -- & $>625$ \\ 
\hline

\textbf{\makecell[l]{CLIPA \\ + DynamiCS (Ours)}} 
& \makecell[c]{LAION-400M \\ (298M)} 
& 1.28B@112 + 128M@224 
& 70.3 & 44.5 & 300 \\ 

\bottomrule
\end{tabular}
}
\end{table}

\section{Conclusion and Outlook}
\label{sec:conclusion}

This paper has proposed a dual-purpose data sampling approach to reduce the training costs of VLMs, important for practical applications, while at the same time balancing the availability of semantic data that is key to performance.
Extensive experiments demonstrate that DynamiCS achieves better full-scale effectiveness at much lower cost while outperforming other low-cost pre-training alternatives. 
Its effectiveness on both LAION-400M and DataComp-DFN suggests it can extend to other imbalanced datasets to reduce training cost.

DynamiCS is based on two established insights.
First, we establish that data sampling should involve both down- and upsampling, to maintain VLMs' performance on the long tail.
The cost savings achieved with dynamic sampling give us the freedom to integrate upsampling into our approach.
The novelty of DynamiCS is that it breaks with the ``aim for even'' philosophy, which cuts off the fat head in an attempt to flatten the distribution, and instead pursues an ``aim for utility'' philosophy.
Second, dynamic sampling makes an important contribution to efficient VLM pre-training: it is very effective at reducing the cost of VLMs, and we expect it to be an element in future methods as well.

\bibliographystyle{unsrt}  
\bibliography{references}  %%% Remove comment to use the external .bib file (using bibtex).

\clearpage
\setcounter{page}{1}
\appendix
\section{Details of Experimental Setup}

We follow OpenCLIP~\cite{gabriel2021openclip}, FLIP~\cite{li2023flip}, and CLIPA~\cite{li2023clipa} to pre-train and evaluate our methods.

\subsection{Architectures}
Following FLIP~\cite{li2023flip} and CLIPA~\cite{li2023clipa}, we used ViT-B/16 and ViT-L/16 with global average pooling as the image encoder.
For models pre-trained on the DataComp and LAION-400M datasets, the input image resolution is 112×112 for both ViT-B/16 (49 image tokens) and for ViT-L/16 (49 image tokens).
During fine-tuning, both models use an input image resolution of $224\times224$.
For the text encoder, we adopted a Transformer model~\cite{vaswani2017Transformer} with byte-pair encoding and a 49K token vocabulary~\cite{gabriel2021openclip}.

\begin{table}[!t]
    \centering
    \caption{Details of the pre-training and fine-tuning setups on \textbf{LAION-400M} and \textbf{DataComp} dataset for Sec.~\ref{sec:results}.
    }
    \label{tab:setup}
    % \resizebox{\linewidth}{!}{%
    \begin{tabular}{l|l|l}
    \toprule
        \textbf{Config} & \textbf{Pre-training} & \textbf{Fine-tuning} \\
        \midrule
        optimizer & AdamW~\cite{loshchilov2017adaW} & AdamW~\cite{loshchilov2017adaW} \\
        batch size & 28k & 8k \\
        base learning rate & 8e-6 & 8e-7 \\
        weight decay & 0.2 & 0.2 \\
        optimizer momentum & $\beta_1, \beta_2=0.9, 0.95$~\cite{chen2020generative} & $\beta_1, \beta_2=0.9, 0.95$~\cite{chen2020generative} \\
        learning rate schedule & cosine decay~\cite{loshchilov2016sgdr} & cosine decay~\cite{loshchilov2016sgdr} \\
        warmup steps & 1600 & 10\% \\
        samples seen & 1.28B & 128M \\
        $\tau$ & 0.07 & --- \\
        numerical precision & amp & amp \\
        RandomResizedCrop &  (40, 100)  & (40, 100)  \\
        \bottomrule
    \end{tabular}
    % }
\end{table}

\subsection{Hyper-parameters}

For our experimental analysis of DynamiCS, we pre-trained and fine-tuned the model on the DataComp dataset. 
For all of the experiments on DataComp and LAION-400M, we follow CLIPA~\cite{li2023clipa} to set the base learning rate of 8e-6 for pre-training and 8e-7 for fine-tuning.
The learning rate is calculated by the linear learning rate scaling rule~\cite{Goyal2017minisgd}: \textit{lr = base\_lr $\times$ batchsize $/~256$}.
The details of the pre-training and fine-tuning procedures are provided in Table~\ref{tab:setup}.

\subsection{Implementation.}

First, we use DINOv2-base~\cite{oquab2024dinov2} (ViT-B/16) to generate the image embeddings of the pre-training datasets.
Then, we employ the K-means clustering algorithm based on cosine similarity (provided by faiss~\cite{douze2024faiss}) to cluster the image embeddings.
Our pre-training codebase is built on OpenCLIP~\cite{gabriel2021openclip}.
And we download the datasets using the img2dataset~\cite{beaumont2021img2dataset} tool.

\subsection{Evaluation Setting}

We evaluate the models using the CLIP Benchmark tool~\cite{gabriel2021openclip,cherti2022clipbench}.

\noindent \textbf{\textit{Let It Wag!} dataset.}
\textit{Let It Wag!} designed by Udandarao et al.~\cite{udandarao2024no}, is designed to evaluate the performance of VLMs on long-tailed concepts.
The dataset contains 290 long-tailed categories selected from 4,029 concepts curated across 27 datasets.
The average concept frequency is about 1,096, and roughly 140 concepts appear fewer than 1,000 times.
The frequency was statistically analyzed using the LAION-400M dataset.
For more details, see~\cite{udandarao2024no}.

\begin{figure*}[!t]
    \centering
    \includegraphics[width=1.0\linewidth]{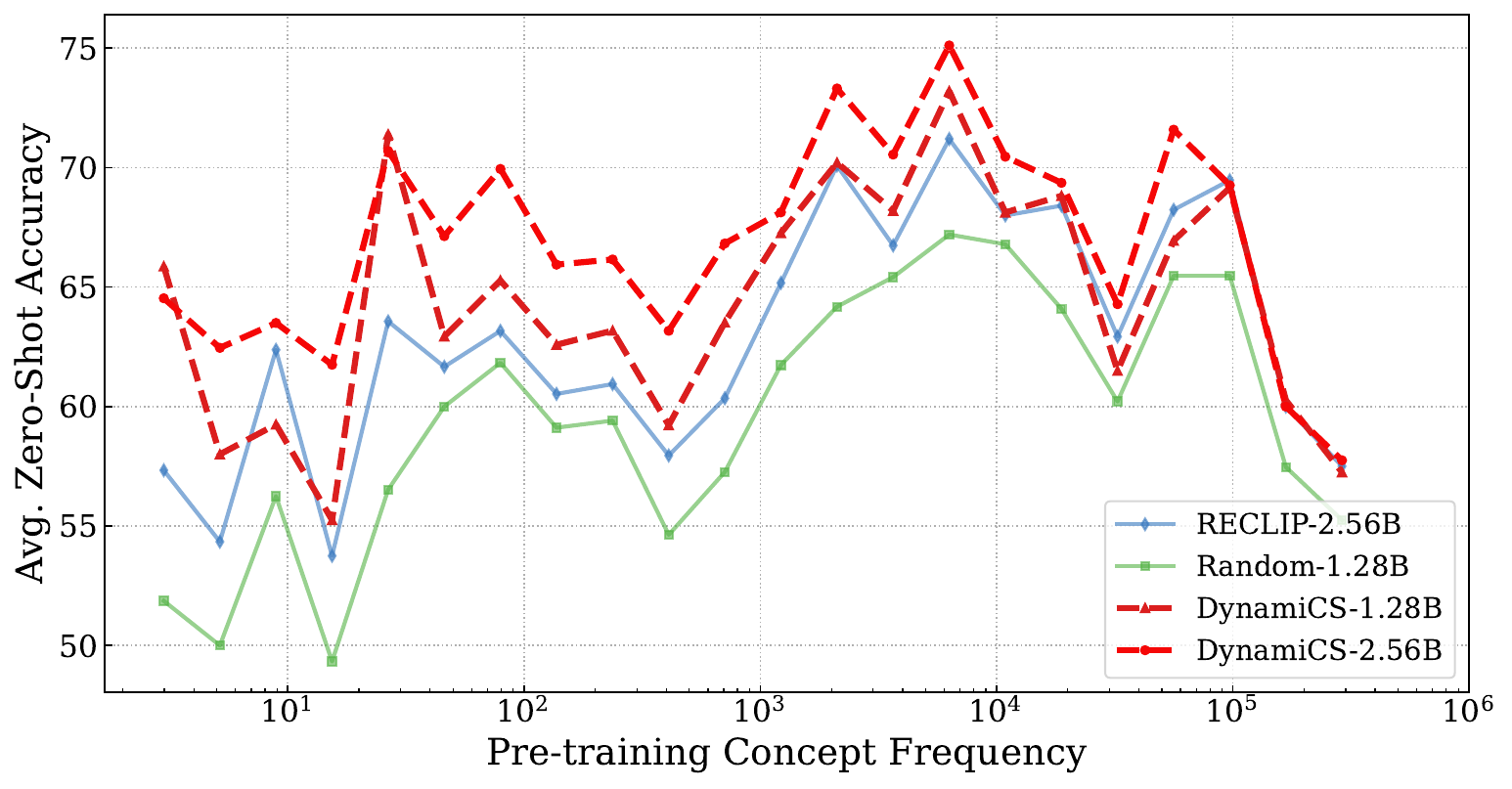}
    \caption{
    \textbf{The Log-linear relationships between concept frequency and zero-shot performance of ImageNet-1K.}
    Our approach outperforms both RECLIP-Random pruning and RECLIP and substantially surpasses them on long-tail categories.
    %All models use the ViT-B/16 image encoder.
    We pre-train RECLIP-Random and RECLIP-DynamiCS with 1.28B samples seen and fine-tune in small steps, reducing their training cost to 50\% of RECLIP.
    Our model was pre-trained on the LAION-400M~\cite{schuhmann2021laion400m} dataset.
    }
    \label{fig:log_linear}
\end{figure*}

\section{More results}

In Section~\ref{sec:results_long_tail}, we evaluate the performance of the models on the long-tail datasets.
We estimate concept frequencies by counting their occurrences in the LAION-400M captions. Unlike~\cite{udandarao2024no}, we use 25 bins, each containing 40 classes, to report the average performance. 
This provides a more fine-grained view of class-level performance.
As shown in Figure~\ref{fig:log_linear}, RECLIP outperforms the RECLIP-Random pruning method across all concept frequencies, since RECLIP sees twice as many samples as the Random pruning method.
DynamiCS also outperforms the RECLIP-random pruning method across all concept frequencies, and DynamiCS-1.28B outperforms the RECLIP method on the long-tail concepts, with only 50\% of samples seen, as DynamiCS up-samples the long-tail concepts.
Meanwhile, DynamiCS shows negligible performance loss on the head (high-frequency) concepts.
Finally, DynamiCS-2.56B further improves performance across both high- and low-frequency concepts.

Interestingly, some points in Figure~\ref{fig:log_linear} do not strictly follow the log-linear relationship.
Even some high-frequency classes show lower performance. We leave a deeper investigation of these cases to future work.

\subsection{Pre-training and Fine-tuning}

\begin{table*}[t!]
\caption{\textbf{Comparison of zero-shot top-1 classification on ImageNet-1K and \textit{Let It Wag!}.}
Our model was pre-trained on the LAION-400M dataset and a subset of \textbf{DataComp-Large} that was filtered by DFN-2B.
FLIP is pre-trained with 75\% image masking, resulting in the same number of image tokens as a 112$\times$112 image size.
CLIPA is pre-trained by syntax masking with 16 text tokens.
The symbol $*$ indicates results we reproduced.
The symbol $\approx$ indicates estimated values, because GPU-hours are not reported in their paper.
All models use the ViT-B/16 image encoder.
}
\label{tab:imagenet_comparison_pre_train}
\footnotesize
\centering
\resizebox{\textwidth}{!}{
\begin{tabular}{l|c|c|cc|cc|c}
\toprule
\textbf{Models} 
& \textbf{\makecell[c]{Dataset \\ (Data Size)}} 
& \textbf{Samples Seen@Resolution}
& \multicolumn{2}{c|}{\textbf{ImageNet-1K}}
& \multicolumn{2}{c|}{\textbf{\textit{Let It Wag!}}}
& \textbf{GPU-hours} \\
\cmidrule(lr){4-5} \cmidrule(lr){6-7}
& & & \textbf{Pre-train} & \textbf{Fine-tune} & \textbf{Pre-train} & \textbf{Fine-tune} & \\
\midrule

\textbf{FLIP~\cite{li2023flip}} 
&  
& 2.56B@224 + 128M@224 
& 58.5 & 60.9 & --- & --- & --- \\
\cmidrule(lr){1-1} \cmidrule(lr){2-8}

\textbf{CLIPA*~\cite{li2023clipa}} 
& \multirow{4}{*}{\makecell[c]{LAION-400M \\ (298M)}} 
& 2.56B@112 + 128M@224 
& 59.8 & 63.2 & 28.3 & 36.4 & 269 \\

\textbf{RECLIP*~\cite{li2023reclip}} 
&  
& 2.56B@112 + 128M@224 
& 59.6 & 62.9 & 29.9 & 36.0 & 280 \\
\cmidrule(lr){1-1} \cmidrule(lr){3-8}

\textbf{DynamiCS (Ours)} 
&  
& 1.28B@112 + 128M@224 
& 60.7 & \textbf{65.0} & 36.3 & 42.1 & 163 \\

\textbf{DynamiCS (Ours)} 
&  
& 2.56B@112 + 128M@224 
& 63.8 & \textbf{67.5} & 38.9 & 45.5 & 299 \\
\midrule

\textbf{DataComp~\cite{gadre2023datacomp}} 
& \multirow{2}{*}{\makecell[c]{DataComp \\ (1.28B)}} 
& 1.28B@224 
& --- & 63.1 & --- & 33.7 & $\approx1070$ \\

\textbf{Captioning} 
&  
& 2.56B@224 
& --- & 59.8 & --- & --- & $\approx2140$ \\
\midrule

\textbf{WhatIf~\cite{li2025whatif}} 
& \makecell[c]{Recap-DataComp-1B \\ (1.4B)} 
& 2.56B@112 + 128M@224 
& --- & 69.2 & --- & --- & --- \\
\midrule

\textbf{DFN~\cite{fang2023dfn}} 
& \multirow{5}{*}{\makecell[c]{DataComp-DFN \\ (130M)}} 
& 1.28B@224 
& --- & 67.8 & --- & --- & $\approx1070$ \\

\textbf{HQ-CLIP~\cite{wei2025hq}} 
& 
& 3.20B@224 
& --- & 70.6 & --- & 38.2 & $\approx2675$ \\

\textbf{DFN*~\cite{fang2023dfn}} 
&  
& 1.28B@112 + 128M@224 
& 64.6 & 68.7 & 36.2 & 42.4 & 151 \\
\cmidrule(lr){1-1} \cmidrule(lr){3-8}

\textbf{DynamiCS (Ours)} 
&  
& 0.64B@112 + 128M@224 
& 64.3 & 69.2 & 40.0 & 46.5 & 95 \\

\textbf{DynamiCS (Ours)} 
&  
& 1.28B@112 + 128M@224 
& 67.5 & \textbf{71.3} & 44.1 & \textbf{50.2} & 163 \\
\bottomrule
\end{tabular}
}

\end{table*}

In Table~\ref{tab:imagenet_other_methods} in Section~\ref{sec:results}, we report only the results after fine-tuning. 
Table~\ref{tab:imagenet_comparison_pre_train} presents the results before fine-tuning. 
As we can see, fine-tuning plays an important role in closing the gap between training and evaluation. 
Even a small amount of fine-tuning (128M samples seen) substantially improves performance on both ImageNet-1K and \textit{Let It Wag!}.

\section{Computing cost}

\textbf{Embeddings} are a one-time offline inference cost and can be reused across experiments.
It takes about 24 and 17 GPU-hours to generate the embeddings for the LAION-400M and DataComp datasets.
In contrast, CLIP-score-based methods, such as DFN and DataComp, require a pre-trained CLIP to generate both image and text embeddings to compute similarity scores for data filtering.

\begin{table}
    \centering
    \caption{GPU-hours required to generate dataset embeddings for LAION-400M and DataComp datasets by DINOv2-ViT-B/16 model.}
    \label{tab:embedding_time}
    \begin{tabular}{l|c}
        Dataset & GPU-hours\\ \hline
        LAION-400M & 23h:42 \\
        DataComp & 16h:58  \\ \bottomrule
    \end{tabular}
\end{table}

\noindent \textbf{Efficient K-means setup.} We perform cosine-similarity K-means clustering on GPU using Faiss~\cite{douze2024faiss}, with \textit{$max\_points\_per\_centroid=1000$} and 10 iterations.
As shown in Table~\ref{tab:cluster_time}, the clustering cost remains low across different numbers of clusters.

\begin{table}[!t]
\caption{GPU-hours required for clustering for the different number of clusters. 
We use \textit{$max\_points\_per\_centroid$}=1000 and run 10 iterations.}
\label{tab:cluster_time}
\footnotesize
\centering
\begin{tabular}{c|c}
\toprule
\multirow{1}{*}{\textbf{Number of Cluster}} & \textbf{GPU-hours} \\ 
\midrule
    10k & 0h:26\\ 
    25k & 2h:41 \\ 
    50k & 7h:10\\ 
    70k & 10h:01\\ 
\bottomrule
\end{tabular}
\end{table}

\noindent \textbf{Preprocessing Overhead}
We have shown the computing cost in Tables~\ref{tab:embedding_time} and~\ref{tab:cluster_time}.
To make it clearer, we provide a full end-to-end breakdown of GPU-hours in Table~\ref{tab:overhead}.
Critically, \textbf{embedding extraction and clustering are one-time offline costs} reused across all experiments and future runs.
As training scales up, preprocessing overhead becomes proportionally smaller.
Moreover, K-means clustering can be made very inexpensive. 
SuperKMeans~\cite{kuffo2026superkmeans} reports up to roughly $100\times$ speedup over FAISS on CPU.

\begin{table}[!t]
\centering
\small
\caption{\textbf{GPU-hour breakdown} of DynamiCS on DataComp. 
Overhead = preprocessing/training.
Preprocessing is once offline.
}
\label{tab:overhead}
\resizebox{\linewidth}{!}{
\begin{tabular}{lcccccc}
\toprule
\textbf{Configuration} & \textbf{Embedding} & \textbf{Clustering} & \textbf{Merging} & \textbf{Training} & \textbf{Total} & \textbf{Overhead} \\
\midrule
DynamiCS (0.64B)    & 17   & 7    & 0.25 & 95  & 119.25 & 25.5\% \\
DynamiCS (1.28B)    & 17   & 7    & 0.25 & 163 & 187.25 & 14.9\% \\
DynamiCS (2.56B)    & 17   & 7    & 0.25 & 299 & 323.25 & 8.1\%  \\
\bottomrule
\end{tabular}
}

\end{table}

\subsection{Pre-training and fine-tuning time}

Table~\ref{tab:imagenet_comparison_1} reports GPU-hour costs for the pre-training stage on DataComp with 112 resolution for different samples seen.
Fine-tuning stage stays constant at 27 GPU-hours across all runs.
Compared to Random (63 pre-train, 90 total), DynamiCS is slightly higher at the same sample budget (68 pre-train, 95 total) and increases to 163 and 299 total GPU-hours as the pre-training samples grow to 1.28B and 2.56B, respectively.

\begin{table*}[t!]
\caption{The GPU-hours for pre-training and fine-tuning stages.}
\label{tab:imagenet_comparison_1}
\footnotesize
\centering
\resizebox{\textwidth}{!}{
\begin{tabular}{l|l|c|ccc}
\toprule
\multirow{2}{*}{\textbf{Models}} & \multirow{2}{*}{\textbf{Dataset}} & \multirow{2}{*}{\textbf{Samples Seen@Resolution}} & \multicolumn{3}{c}{\textbf{GPU-hours}}  \\
& & & pre-train & fine-tune & total \\
\midrule
\textbf{Random} &  \multirow{4}{*}{\textbf{\textbf{DataComp}}}  & 0.64B@112 + 128M@224 & 63 & 27 & 90  \\ \cline{0-0} \cline{3-6}
\multirow{2}{*}{\textbf{\textbf{DynamiCS (Ours)}}}  
& & 0.64B@112 + 128M@224 & 68  & 27 & 95 \\
& & 1.28B@112 + 128M@224 & 136 & 27 & 163 \\
& & 2.56B@112 + 128M@224 & 272 & 27 & 299 \\
\bottomrule
\end{tabular}
}
\end{table*}

\section{Ablation}

\subsection{Post-clustering Cluster Refinement}

After clustering the embeddings, some clusters remain semantically redundant because their centroids are close to each other in the embedding space. 
To reduce this redundancy, we apply a greedy semantic fusion procedure. Specifically, we consider cluster centroids one by one and merge a centroid into an existing semantic group if its maximum cosine similarity to the retained centroids exceeds a threshold; otherwise, it starts a new semantic group. 
In this way, the threshold controls how aggressively we fuse nearby semantic concepts.
Table~\ref{tab:cluster_refine} reports ImageNet-1K zero-shot accuracy under different fusion thresholds.

\begin{table}[!t]
    \centering
    \caption{\textbf{Zero-shot classification on ImageNet-1K for different cluster deduplicate}.
    All models are pre-trained on \textbf{DataComp} for 106 million samples seen with ViT-B/16 image encoder and $112\times112$ image resolution.
}
    \label{tab:cluster_refine}
    \begin{tabular}{c|ccccccc}
    \toprule
     & \textbf{0.6} & \textbf{0.7$^\star$} & \textbf{0.75} & \textbf{0.8} & \textbf{0.85} & \textbf{0.9} & \textbf{1.0} \\
    \midrule
    ImageNet-1K        & 38.9  & \textbf{39.2} & 38.8  & 37.7  & 36.8  & 35.6  & 34.9  \\
    Centroids Retained & 5,632 & 10,808        & 15,049 & 21,277 & 29,581 & 39,300 & 50,000 \\
    \bottomrule
    \end{tabular}
\end{table}

\subsection{The number of Cluster}

\begin{table}[!t]
\caption{Zero-shot classification on ImageNet-1K and \textit{Let It Wag!} across \textbf{different number of cluster.}
All models are pre-trained on \textbf{DataComp} for 106 million samples seen with ViT-B/16 image encoder and $112\times112$ image resolution.
}
\label{tab:cluster}
\footnotesize
\centering
\begin{tabular}{l|c|c|c}
\toprule
\multirow{1}{*}{\textbf{Models}} & \multirow{1}{*}{\textbf{Number of Cluster}} & \textbf{ImageNet-1K} & \textbf{\textit{Let It Wag!}}   \\ 
\midrule
\multirow{4}{*}{\textbf{DynamiCS~}} 
& 10k & 39.0  & 19.4\\ 
& 25k & 39.1 & 19.2 \\ 
& 50k & 39.2 & 20.2 \\ 
& 70k & 39.0 & 19.0  \\ 
\bottomrule
\end{tabular}
\end{table}

Table \ref{tab:cluster} studies how the number of clusters used by DynamiCS affects zero-shot accuracy on ImageNet-1K and the long-tail \textit{Let It Wag!} datasets. 
Overall, performance is stable across a wide range of cluster counts: ImageNet-1K remains around 39.0–39.2\% and \textit{Let It Wag!} around 19.0–20.2\%. 

\subsection{The fine-tune batch size}

\begin{table}[!t]
\caption{Zero-shot classification on ImageNet-1K and \textit{Let It Wag!} across different \textbf{fine-tuning batch size} for 1.28B samples seen. The model is pre-trained on the DataComp dataset.
}
\label{tab:batch_size}
\footnotesize
\centering
\begin{tabular}{l|c|c|c}
\toprule
\multirow{1}{*}{\textbf{Models}} & \multirow{1}{*}{\textbf{batch size}} & \textbf{ImageNet-1K} & \textbf{\textit{Let It Wag!}}   \\ 
\midrule
\multirow{4}{*}{\textbf{DynamiCS-1.28B}} 
% & 2k &  & \\ 
& 4k  & 71.1 & 50.1 \\ 
& 8k  & 71.3 & 50.2 \\ 
& 16k & 71.6 & 50.5 \\ 
& 32k & 71.7 & 51.6 \\ 
\bottomrule
\end{tabular}
\end{table}

CLIP training with contrastive learning, which benefits from a large batch size.
Therefore, we study the effect of batch size on zero-shot classification performance for DynamiCS-1.28B with 1.28B samples seen, pre-trained on the DataComp dataset. 
As shown in Table~\ref{tab:batch_size}, increasing the fine-tuning batch size consistently improves performance on both ImageNet-1K and \textit{Let It Wag!}. Specifically, ImageNet-1K accuracy rises from 71.1\% at a batch size of 4k to 71.7\% at 32k, while \textit{Let It Wag!} improves from 50.1\% to 51.6\%.

\end{document}